\documentclass[10pt,twocolumn,letterpaper]{article}

\usepackage{iccv}
\usepackage{times}
\usepackage{epsfig}
\usepackage{graphicx}
\usepackage{amsmath}
\usepackage{amssymb}
\usepackage{subfig}
\usepackage{booktabs}
\usepackage{threeparttable}

\usepackage{tablefootnote}

\usepackage[pagebackref=true,breaklinks=true,letterpaper=true,colorlinks,bookmarks=false]{hyperref}

\iccvfinalcopy 

\def\iccvPaperID{9} 

\begin{document}

\title{Multi-view PointNet for 3D Scene Understanding}

\author{Maximilian Jaritz\textsuperscript{1*}, Jiayuan Gu\textsuperscript{$\dagger$}, Hao Su\textsuperscript{$\dagger$}\\
Inria/Valeo\textsuperscript{*}, UC San Diego\textsuperscript{$\dagger$}\\
{\tt\small maximilian.jaritz@inria.fr, \{jigu,haosu\}@eng.ucsd.edu}
}
\maketitle

\begin{abstract}
   Fusion of 2D images and 3D point clouds is important because information from dense images can enhance sparse point clouds. However, fusion is challenging because 2D and 3D data live in different spaces. In this work, we propose MVPNet (Multi-View PointNet), where we aggregate 2D multi-view image features into 3D point clouds, and then use a point based network to fuse the features in 3D canonical space to predict 3D semantic labels. To this end, we introduce view selection along with a 2D-3D feature aggregation module. Extensive experiments show the benefit of leveraging features from dense images and reveal superior robustness to varying point cloud density compared to 3D-only methods. On the ScanNetV2~\cite{dai2017scannet} benchmark, our MVPNet significantly outperforms prior point cloud based approaches on the task of 3D Semantic Segmentation. It is much faster to train than the large networks of the sparse voxel approach~\cite{graham20183d}. We provide solid ablation studies to ease the future design of 2D-3D fusion methods and their extension to other tasks, as we showcase for 3D instance segmentation.
\end{abstract}

\stepcounter{footnote}\footnotetext{Research done while visiting UC San Diego.}
\section{Introduction}
The field of 3D perception is evolving at a fast pace, with recent major improvements on tasks such as semantic segmentation and object detection. This is crucial to applications in robotics and AR/VR, where 3D data are typically captured as depth maps or point clouds, along with 2D images from RGB cameras. A central problem of those applications is how we can efficiently fuse data from the 2D and 3D domains. This is quite challenging, because there usually is no one-to-one mapping between 2D and 3D data, and also the neighborhood definitions in 2D and 3D are different for convolution. More critically, while neighboring pixels are defined by the discrete grid, 3D points are defined at non-uniform continuous locations. Additionally, 3D sensors mostly deliver a much lower resolution than 2D cameras. For example, when the point cloud from a Velodyne HDL-64 Lidar is projected into the camera image, it covers only 5.9\% of the pixels~\cite{jaritz2018sparse}.

\textit{Point cloud based neural networks} have been shown to generate powerful geometry cues for 3D scene understanding. However, not all objects can be distinguished by their shape, especially when they have flat surfaces such as doors, refrigerators and curtains. Therefore, additional color information should be leveraged, but recent results~\cite{wu2018pointconv} have shown that naively feeding colored point cloud (XYZRGB) to \textit{point cloud based networks} does only marginally improve the performance over simple point cloud input (XYZ).

\begin{figure}
	\centering
	\includegraphics[width=\linewidth]{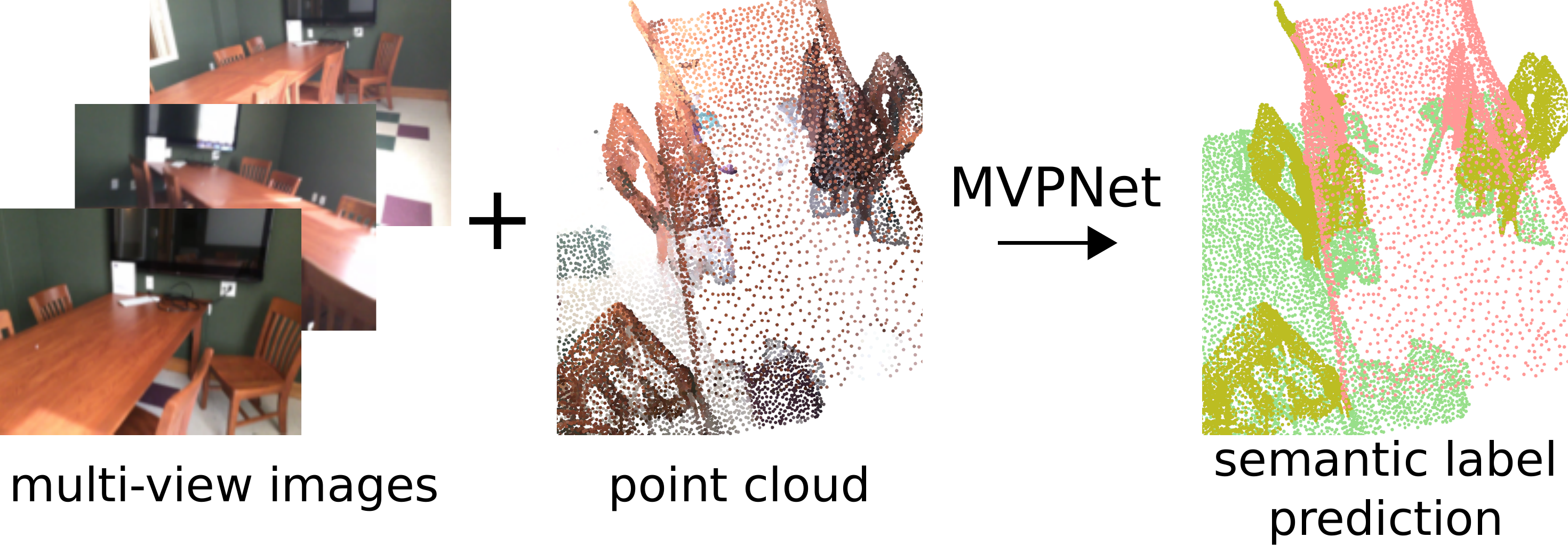}
	\caption{Our MVPNet (Multi-View PoinNet) takes dense multi-view images and a sparse point cloud as input and fuses them to predict the semantic labels for each point.}
	\label{fig:teaser}
\end{figure}

We argue that, because RGB cameras have much higher spatial resolution than 3D sensors in most realistic settings, it is better to compute image features in 2D first before lifting the 2D information to 3D. Like so, it is possible to gather \textit{additional information} from higher resolution images and it is also \textit{natural} from a sensor fusion perspective, to push modality centric features from different sources to 3D for their combination.

As different representations in 3D exist (voxel, point-cloud, multi-view, etc.), their respective scene understanding methods evolve in parallel. For voxel-based methods, there have been works on how to fuse geometry and image data coherently. However, for the point cloud domain, the common practice is to sparsely copy RGB information to points and there lacks a systematic exploration of how to conduct the fusion more effectively. In order to address this significant drawback of point cloud based methods, we propose MVPNet (Multi-View PointNet), where we first compute 2D image features on multiple, heuristically selected frames, then lift those features to 3D and adaptively aggregate them into the original point cloud (XYZ). Finally, the multi-view augmented point cloud is fed into PointNet++~\cite{qi2017pointnet++} for semantic segmentation.

There are several advantages: First, the lifted 2D features contain contextual information thanks to the receptive field of the 2D network. Second, the complementary RGB and geometry features are jointly processed in canonical 3D space. And third, our flexible approach can be added to any 3D network.

In this paper, we focus on exploring the 2D-3D fusion problem, a key component for 3D scene parsing.
While the key message of our exploration can be concisely summarized as doing early feature fusion is better, the significant performance improvement from baselines is in fact obtained through extensive trials. In the experiment section, we made a rich set of ablation studies so as to compare design choices and inform our discoveries to the vision community.



In summary, the key contributions are as follows:
\begin{itemize}
    \item We propose a simple and fast framework that takes 3D point cloud and 2D RGB-D frames as input and fuses complementary features in the \textit{canonical point cloud space} for the task of 3D semantic segmentation.
    \item Our method outperforms previously published \textit{point cloud based networks} by using additional dense image information while handling occlusions.
    \item We provide insights to the design choices in \text{dense-2D/sparse-3D point cloud fusion} based on extensive experiments, and showcase its excellent robustness to very sparse point clouds.
\end{itemize}

\section{Related Work}
\paragraph{2D to 3D Lifting}
Several works have shown that lifting 2D features to 3D leads to better performance than just lifting RGB values. In~\cite{dai20183dmv}, multiple 2D image feature maps are unprojected to 3D, voxel-volumes are created, combined by max-pooling and then fed into a 3D CNN.
In~\cite{liang2018deep}, 2D image features are gathered at nearest neighbor locations defined by a lidar point cloud to build a dense bird view map.
These approaches use pixel-level 2D-3D correspondences to lift low-level features as opposed to \cite{qi2018frustum} where only high-level 2D object proposals are lifted to 3D frustums.
In this work, we establish pixel-to-point correspondences to lift 2D features to the canonical 3D point cloud space instead of voxel~\cite{dai20183dmv} or birdview~\cite{liang2018deep}. The advantage is that once all modalities are represented in a 3D point cloud, correspondence between two data points is precisely defined by distance in the continuous domain without discretization errors.

\paragraph{3D Networks}
CNNs are the state-of-the-art on 2D RGB images, but competing network families exist for 3D data: 3D CNNs~\cite{maturana2015voxnet, ji20133d} make use of the voxel representation where the raw point cloud data is transformed into a discrete grid of cells and in practice most of the cells are empty and only voxels that lie on the object surface are occupied. On the other hand, \textit{point cloud based networks}~\cite{qi2017pointnet, klokov2017escape, wang2018dynamic, li2018pointcnn, wu2018pointconv, xu2018spidercnn} can directly take point clouds as input. In our work, we use point cloud based networks, because of their inherent sparsity as compared to voxel-based methods.

\paragraph{3D Semantic Segmentation}
The aim of 3D semantic segmentation is to predict a label for every point in a 3D point cloud.
PointNet~\cite{qi2017pointnet} leverages shared Multi Layer Perceptrons (MLPs) to compute point-wise features and uses max-pooling to obtain features for the global point cloud. 
This works very well for single objects in the ShapeNet dataset~\cite{chang2015shapenet} for the task of part segmentation.
For whole scene analysis, PointNet++~\cite{qi2017pointnet++} is more suited, because it has set abstraction layers to create a hierarchical network structure akin to CNNs which scales much better to larger point sets.
Voxel-based methods include SegCloud~\cite{tchapmi2017segcloud}, 3DMV~\cite{dai20183dmv} and Submanifold Sparse Convolution~\cite{graham20183d}.
The latter defines a very efficient way to deal with sparsely populated voxels by restricting computations to active voxels.
Different with 3DMV~\cite{dai20183dmv}, we exploit the fusion of multi-view and geometry information in point cloud space and achieve much better performance. In addition, we report the mIoU for all the ablation studies instead of the segmentation accuracy.
SPLATNet~\cite{su2018splatnet} takes point clouds and images as input and projects them on a permutohedral lattice for convolution and 2D-3D fusion. In our approach, we focus on fusing multi-view features with an aggregation module directly in the canonical point cloud space and achieve higher mIoU (64.1) than SPLATNet (39.3) on the ScanNet benchmark.

\paragraph{3D Instance Segmentation}
The task of 3D instance segmentation is more precise than 3D object detection: Instead of regressing boxes, point masks which describe the exact shape of each object are predicted.
Proposal based approaches like Mask R-CNN\cite{he2017mask} are the state-of-the-art in 2D and have been extended to 3D by leveraging voxels and 3D box-proposals~\cite{hou20183d}. Alternatively to proposing boxes, point clouds can be generated as proposals~\cite{yi2018gspn}.
Another strategy is clustering based on predicted semantic labels or a similarity matrix~\cite{wang2018sgpn} which can be learned~\cite{liu2019masc}.
We extend MVPNet to instance segmentation using R-PointNet~\cite{yi2018gspn}.

\section{MVPNet}\label{sec:method}
\begin{figure*}
	\centering
	\includegraphics[width=0.97\textwidth]{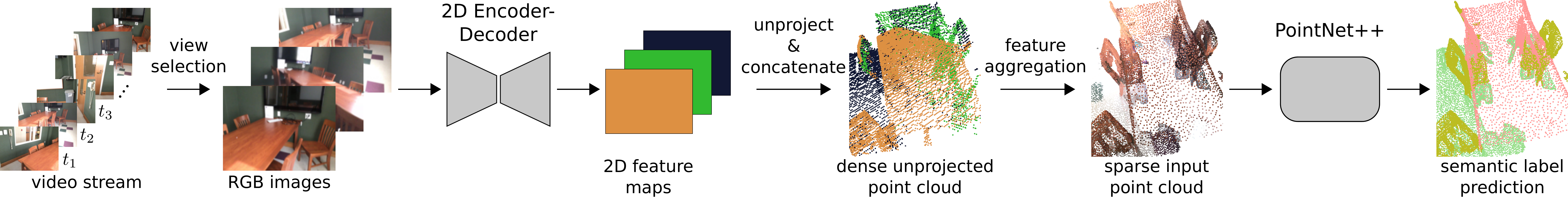}
	\caption{Pipeline overview. First, a fixed number of 2D views are selected so that the whole 3D scene is maximally covered. Then, the respective RGB images are fed into a 2D encoder-decoder to obtain feature maps of same size as the input images. Those feature maps are unprojected and concatenated to form a dense point cloud. Then, the dense unprojected feature point cloud is aggregated into the sparse input point cloud to augment each point with 2D image features. Complementary 3D geometry and 2D image features are fused in 3D canonical space using PointNet++ which predicts the final semantic labels.}
	\label{fig:pipeline}
\end{figure*}

Our MVPNet is designed to effectively fuse complementary information from multiple RGB-D frames and 3D point cloud in order to achieve better 3D scene understanding on real-world data, like ScanNetV2~\cite{dai2017scannet}.
The primary task is 3D semantic segmentation, where the goal is to predict a semantic label for each point in the input point cloud. Our pipeline is illustrated in Fig.~\ref{fig:pipeline}. We also showcase an extension to 3D instance segmentation in Sec.~\ref{sec:3dInstSeg}.

\subsection{Overview}
The data of each scene consists of a sequence of RGB-D frames and a point cloud. The input point cloud, denoted as $\mathcal{S}_{\text{sparse}}$, is sparse compared with the resolution of images. This can be seen in Fig.~\ref{fig:frameSelection} by comparing the density of the sparse point cloud with the unprojected views.
Following PointNet++~\cite{qi2017pointnet++}, we divide the whole scene into chunks (around 90 chunks for an average scene).
For each chunk, the most $M$ informative views (RGB-D frames) are selected to maximize the coverage of the input point cloud (Sec.~\ref{sec:viewSelection}). 
Those views (RGB) are then fed into a 2D encoder-decoder network in order to compute $M$ feature maps (Sec.~\ref{sec:2dEncoderDecoder}).
To augment the sparse input point cloud $\mathcal{S}_{\text{sparse}}$, pixels with valid depth in each 2D feature map are first lifted to a 3D point cloud and then a dense point cloud $\mathcal{S}_{\text{dense}}$ is obtained by concatenating all the $M$ unprojected point clouds.
Given the image features associated with $\mathcal{S}_{\text{dense}}$, our feature aggregation module samples the $k$ nearest neighboring points in $\mathcal{S}_{\text{dense}}$ and adaptively combines them to form the new feature for the point in $\mathcal{S}_{\text{sparse}}$ (Sec.~\ref{sec:2d3dFeatureLift}). 
Finally, we leverage PointNet++ to process the \textit{multi-view feature augmented point cloud} from a 3D geometric perspective.

\subsection{View Selection}\label{sec:viewSelection}
\begin{figure}
	\centering
	\newcommand\len{0.23}
	\subfloat[sparse PC]{
		\includegraphics[width=\len\linewidth]{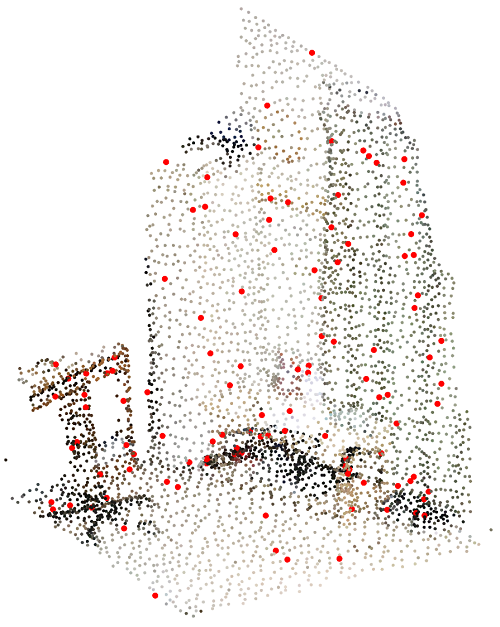}
		\label{fig:frameSelect0}
	}
	\subfloat[1st view]{
		\includegraphics[width=\len\linewidth]{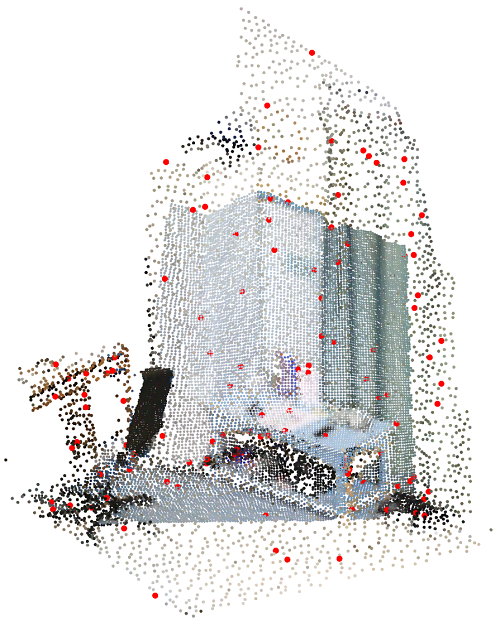}
		\label{fig:frameSelect1}
	}
	\subfloat[2nd view]{
		\includegraphics[width=\len\linewidth]{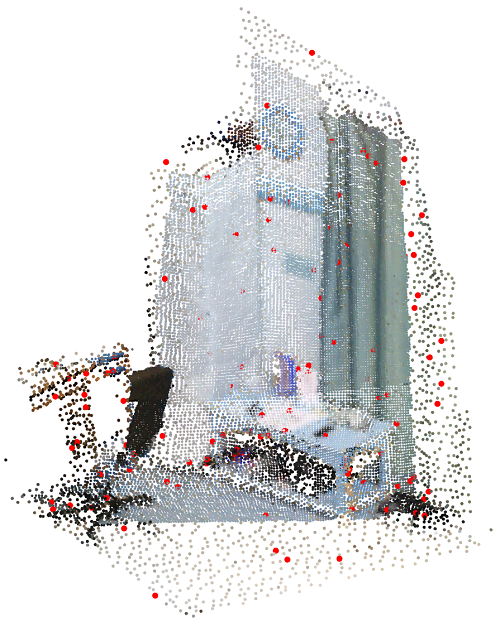}
		\label{fig:frameSelect2}
	}
	\subfloat[3rd view]{
		\includegraphics[width=\len\linewidth]{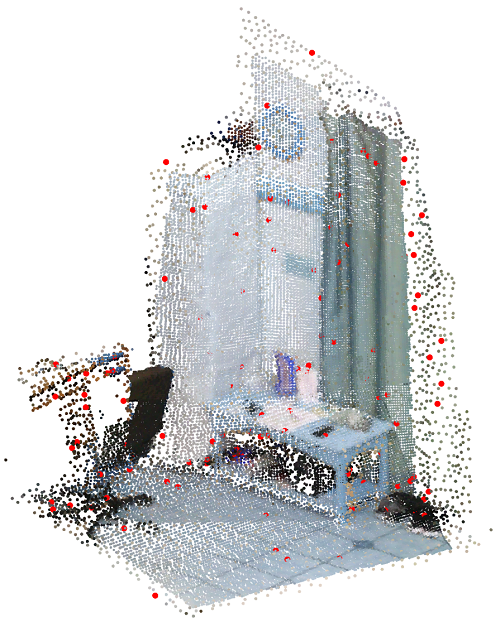}
		\label{fig:frameSelect3}
	}
	\caption{View selection:
		\ref{fig:frameSelect0} visualizes the input point cloud of a chunk  and its coarse version (red points) used to compute the overlap with the RGB-D frames.
		\ref{fig:frameSelect1}, \ref{fig:frameSelect2} and \ref{fig:frameSelect3} show the 1st, 2nd and 3rd greedily selected view.}
	\label{fig:frameSelection}
\end{figure}
In ScanNetV2~\cite{dai2017scannet}, the RGB-D frames come as video stream with strong overlap between consecutive frames. It would be redundant and computationally expensive to process them all. Therefore, we make a selection of 1 to 5 views, which maximize contained information, to fuse with the point cloud of the scene. 

In the preprocessing step, the overlaps between the scene point cloud and all the unprojected RGB-D frames of the video stream are computed. To reduce computation, we downsample the point cloud (red points in Fig.~\ref{fig:frameSelection}).
During training we use the overlap information to select the RGB-D frames on-the-fly with a greedy algorithm. The image which overlaps with the most yet uncovered points is selected. We found that this straightforward but efficient method can achieve very high coverage even with few frames, leading to better results with same computation.

\subsection{2D Encoder-Decoder Network}\label{sec:2dEncoderDecoder}
We feed the selected RGB images into a 2D encoder-decoder network based on U-Net~\cite{ronneberger2015u} to compute image feature maps.
In our implementation, the size of the input image is equal to that of the output feature map, and fixed to $160\times120$.
With the relatively low resolution, we found UNet to be better suited in terms of memory, speed, and performance than other 2D semantic segmentation architectures such as DeepLabv3~\cite{chen2017rethinking}, PSPNet~\cite{zhao2017pyramid}, optimized for a much higher resolution.
We pretrain the 2D encoder-decoder network on the task of 2D segmentation on ScanNetV2 in order to bootstrap the training of the whole pipeline.
More details can be found in Sec.~\ref{sec:impl}.

\subsection{2D-3D Feature Lifting Module}\label{sec:2d3dFeatureLift}
In order to obtain the 3D coordinates for the feature maps that have been computed with the RGB images and the 2D encoder-decoder network, we unproject the corresponding depth maps using the camera instrinsics and poses.
Consider $M$ 2D feature maps of size $H \times W \times C_{\text{feat}}$, then each one is lifted to a point cloud of size $N_{\text{RGB}} \times C_{\text{feat}}$, where $N_{\text{RGB}} < HW$ is a hyperparameter that corresponds to the number of unprojected pixels in each RGB image. By concatenating all the $M$ unprojected points together, we yield a dense point cloud $\mathcal{S}_{\text{dense}}$ of size $MN_{\text{RGB}} \times C_{\text{feat}}$.

For semantic segmentation, the labels have to be predicted for the input point cloud $\mathcal{S}_{\text{sparse}}$.
Thus, we have to transfer the features from the unprojected point cloud $\mathcal{S}_{\text{dense}}$ to $\mathcal{S}_{\text{sparse}}$.
Therefore, we use our feature aggregation module which includes a shared MLP inspired by \cite{liang2018deep} in order to distill a new feature for each point in $\mathcal{S}_{\text{sparse}}$ from its $k$ nearest neighbors in $\mathcal{S}_{\text{dense}}$
\begin{align}\label{eq:featAggreg}
\textbf{h}_i = \sum_{j \in \mathcal{N}_k(i)} \text{MLP}(\text{concat}\,[\textbf{f}_j, \textbf{f}_{\text{dist}}\left(\textbf{x}_i, \textbf{x}_j\right)]
\end{align}
where $\textbf{h}_i$ is the distilled feature at point $\textbf{x}_i$ in $\mathcal{S}_{\text{sparse}}$, $\textbf{f}_j$ the semantic feature at one of the $k$ nearest neighbors points $\textbf{x}_j$ in $\mathcal{S}_{\text{dense}}$, and $\textbf{f}_{\text{dist}}(\textbf{x}_i, \textbf{x}_j)$ the distance feature between the two points which we define as
\begin{align}\label{eq:distFeat}
\textbf{f}_{\text{dist}}\left(\textbf{x}_i, \textbf{x}_j\right) = \text{concat}\,[\textbf{x}_i - \textbf{x}_j, \left\lVert \textbf{x}_i - \textbf{x}_j\right\rVert^2].
\end{align}
We define \textit{multi-view feature augmented point cloud} as the resulting features associated with 3D coordinates.
Note that the whole 2D-3D feature lifting module is differentiable, which enables end-to-end training of our MVPNet.

\subsection{3D Fusion Network}\label{sec:3dFusion}
\begin{figure}
	\centering
	\includegraphics[width=0.9\linewidth]{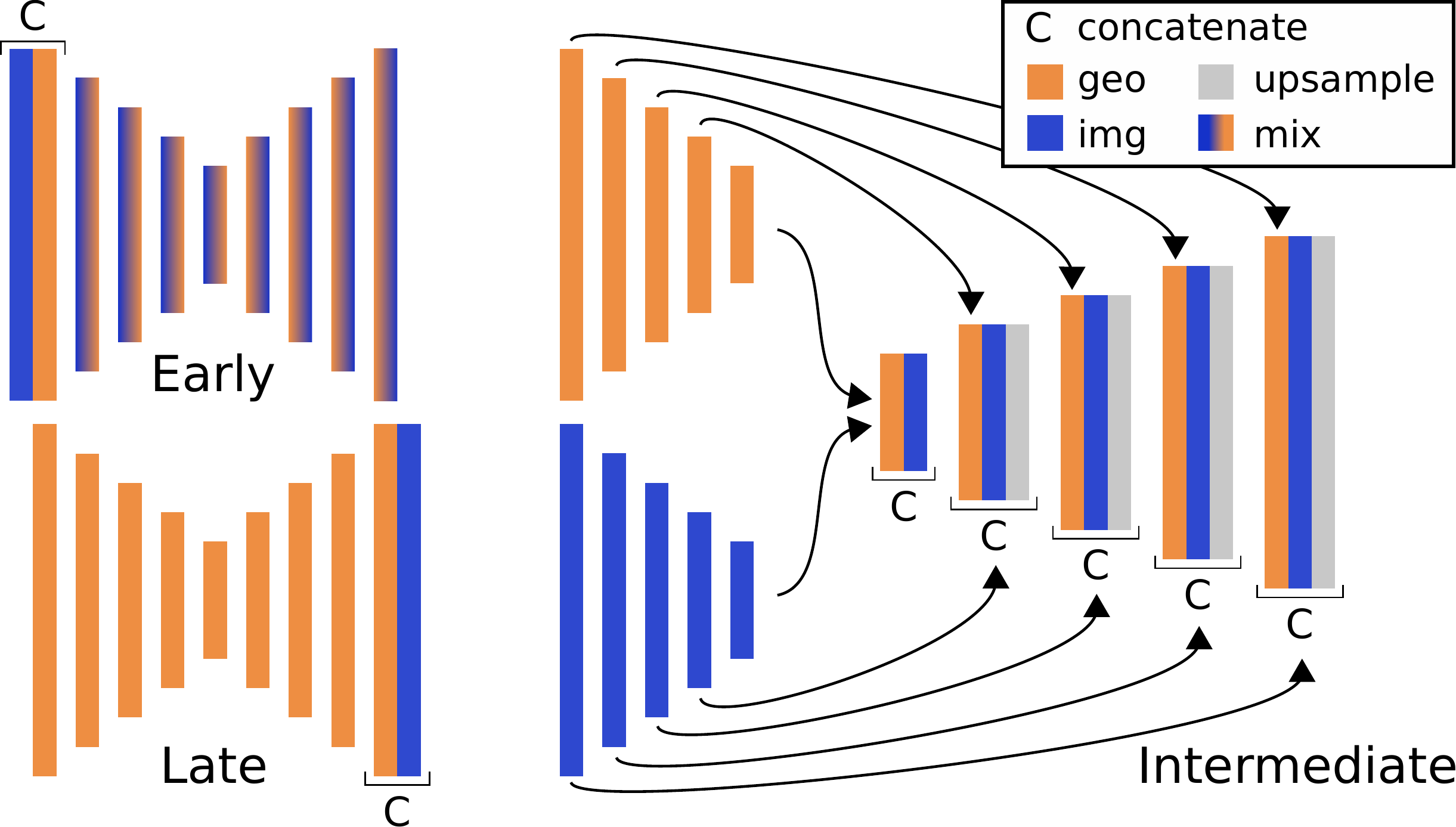}
    \caption{Fusion architectures based on PointNet++. We adopt early fusion, where the geometry (XYZ) and image features are concatenated at the input layer. The network fuses them which leads to mixed features for the remaining network. Other strategies are described in Sec.~\ref{sec:3dFusion}.}
	\label{fig:fusion}
\end{figure}
To fuse multi-view image features and geometry information, we employ PointNet++~\cite{qi2017pointnet++} as backbone.
The original PointNet++ consumes both the coordinates and its corresponding features, such as normal or color. For 3D semantic segmentation, it encodes the input point cloud with set abstraction layers hierarchically, and decodes the output semantic prediction through feature propagation layers. The 3D coordinates of input points are concatenated to the output features of each set abstraction layer.

We adopt \textit{early fusion}, where the image features are concatenated to the geometry (XYZ) and then given as input to PointNet++. Thus, the network is able to fully exploit the image features from a geometric perspective.
We also investigated \textit{intermediate fusion} and \textit{late fusion}. 
In \textit{late fusion}, the image features are concatenated after the final feature propagation layer in PointNet++, right before the segmentation head.
In \textit{intermediate fusion}, we introduce separate encoder branches for geometry and image features whose outputs are then concatenated and fed into the decoder. Additionally, the decoder leverages the intermediate outputs of two encoder branches through skip connections. The different fusion strategies are illustrated in Fig.~\ref{fig:fusion}.

\section{Experiments}
In this section we cover experiments on the ScanNetV2~\cite{dai2017scannet} dataset, but additional results on S3DIS~\cite{armeni2017joint} can be found in the supplementary where we improve over previous methods by 4.16 mIoU.
\subsection{ScanNetV2 Dataset}
The ScanNetV2 dataset~\cite{dai2017scannet} features indoor scenes like offices and living rooms for which a total of 2.5M frames were captured with the internal camera of an IPad and an additionally mounted depth camera.
The data for each scan consists of an RGB-D sequence with associated poses, a whole scene mesh, as well as semantic and instance labels.
There are 1201 training and 312 validation scans that were taken in 706 different scenes, thus each scene was captured about 1 to 3 times.
The test set contains 100 scans with hidden ground truth, used for the benchmark.

\subsection{Implementation Details}\label{sec:impl}
For the task of 3D semantic segmentation, we follow the same chunk-wise pipeline as PointNet++~\cite{qi2017pointnet++}. During training, one chunk ($1.5m \times 1.5m$ in xy-plane, parallel to ground surface) is randomly selected from the whole scene if it contains more than 30\% annotated points. Random rotation along the up-axis is applied for data augmentation. During testing, the network predicts all the chunks with a stride of $0.5m$ in a sliding-window fashion through the xy-plane. A majority vote is conducted for the points that have predictions from multiple chunks.

We downsample the images and depth maps to a resolution of $160 \times 120$.
Random horizontal flip is applied to augment images during training. We fix the number of unprojected points per RGB-D frame to 8192 of a total 19200 pixels for a resolution of $160 \times 120$. Note that even though many pixels are not lifted to 3D, they are still essential for the 2D feature computation as they lie in the receptive field of unprojected pixels.

The backbone of the 2D Encoder network is an ImageNet-pretrained VGG16~\cite{simonyan2014very} with batch normalization and dropout.
For ablation studies and submissions, we also experiment with VGG19~\cite{simonyan2014very} and ResNet34~\cite{he2016deep}.
The custom 2D Decoder network is a lightweight variant of U-Net~\cite{ronneberger2015u}.
Batch normalization and ReLU are added after each convolution layer in the decoder.

For each chunk, 8192 points are sampled from the input point cloud and augmented by the views selected by the method described in Sec.~\ref{sec:viewSelection}. For the feature aggregation module, we use a two-layer MLP with 128 and 64 channels. To predict the semantic labels for the \textit{multi-view feature augmented point cloud}, we use PointNet++ with single-scale grouping (SSG) as our 3D backbone. However, note that our MVPNet can adapt to any 3D network.

Each epoch consists of 20000 randomly sampled chunks, and the batch size of chunks is 6. The network is trained with the SGD optimizer for 100 epochs. We use a weight decay of 0.0001 and a momentum of 0.9. The learning rate is 0.01 for the first 60 epochs, and then divided by 10 every 20 epochs. MVPNet
is trained on a single GTX 1080Ti.

\subsection{Results for 3D Semantic Segmentation}
\begin{table*}
	\addtolength{\tabcolsep}{-3pt}
	\scriptsize
	\centering
	\begin{threeparttable}
	\begin{tabular}{lc|cccccccccccccccccccc}
		\toprule
		Method & mIoU & bath & bed & bkshf & cab & chair & cntr & curt & desk & door & floor & other & pic & fridge & shower & sink & sofa & table & toilet & wall & window \\
		\midrule
		PointNet++\cite{qi2017pointnet++} & 33.9 & 58.4 & 47.8 & 45.8 & 25.6 & 36.0 & 25.0 & 24.7 & 27.8 & 26.1 & 67.7 & 18.3 & 11.7 & 21.2 & 14.5 & 36.4 & 34.6 & 23.2 & 54.8 & 52.3 & 25.2\\
		Re-impl. PointNet++\tnote{*} & 44.2 & 54.8 & 54.8 & 59.7 & 36.3 & 62.8 & 30.0 & 29.2 & 37.4 & 30.7 & 88.1 & 26.8 & 18.6 & 23.8 & 20.4 & 40.7 & 50.6 & 44.9 & 66.7 & 62.0 & 46.2 \\
		PointCNN\cite{li2018pointcnn} & 45.8 & 57.7 &	61.1 &	35.6 &	32.1 &	71.5 &	29.9 &	37.6 &	32.8 &	31.9 &	\textbf{94.4} &	28.5 &	16.4 &	21.6 &	22.9 &	48.4 &	54.5 &	45.6 &	75.5 &	70.9 &	47.5 \\
		PointConv\cite{wu2018pointconv} & 55.6 & 63.6 &	64.0 &	57.4 &	47.2 &	73.9 &	\textbf{43.0} &	43.3 &	41.8 &	44.5 &	\textbf{94.4} &	37.2 &	18.5 &	46.4 &	\textbf{57.5} &	54.0 &	63.9 &	50.5 &	82.7 &	76.2 &	51.5 \\
		\midrule
        Ours & \textbf{64.1} & \textbf{83.1} & \textbf{71.5} & \textbf{67.1} & \textbf{59.0} & \textbf{78.1} & 39.4 & \textbf{67.9} & \textbf{64.2} & \textbf{55.3} & 93.7 & \textbf{46.2} & \textbf{25.6} & \textbf{64.9} & 40.6 & \textbf{62.6} & \textbf{69.1} & \textbf{66.6} & \textbf{87.7} & \textbf{79.2} & \textbf{60.8} \\
		\bottomrule
	\end{tabular}
	\begin{tablenotes}
	\item[*] Anonymous third-party submission. Included for fair comparison because the original PointNet++ results seem very low and are inconsistent with our experiments.
	\end{tablenotes}
	\end{threeparttable}
	\caption{Comparison with the state-of-art \textit{point cloud based methods} on ScanNetV2 3D Semantic label benchmark.}
	\label{tab:resultsSegBenchmark-Point}
\end{table*}

\begin{table}
	\scriptsize
	\centering
	\begin{tabular}{lc}
		\toprule
		Method & mIoU\\
		\midrule
		SparseConvNet\cite{graham20183d} & \textbf{72.5}\\
		3DMV\cite{dai20183dmv} & 48.4\\
		\midrule
		Ours & 64.1\\
		\bottomrule
	\end{tabular}
	\caption{Comparison with \textit{voxel based methods} on the ScanNetV2 3D Semantic label benchmark.}
	\label{tab:resultsSegBenchmark-Voxel}
\end{table}

We evaluate our MVPNet on the ScanNetV2 3D semantic label benchmark. The evaluation metric is the average IoU (mIoU) over 20 classes.
For submission, we ensemble 4 models of MVPNet, which consumes 5 views and uses ResNet34 as 2D backbone.

\begin{figure*}
	\centering
	\newcommand\len{0.18}
	
	\subfloat{
		\includegraphics[width=.84\linewidth]{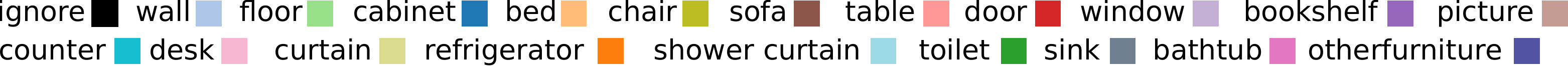}
	}
	
	\subfloat{
		\includegraphics[width=\len\linewidth]{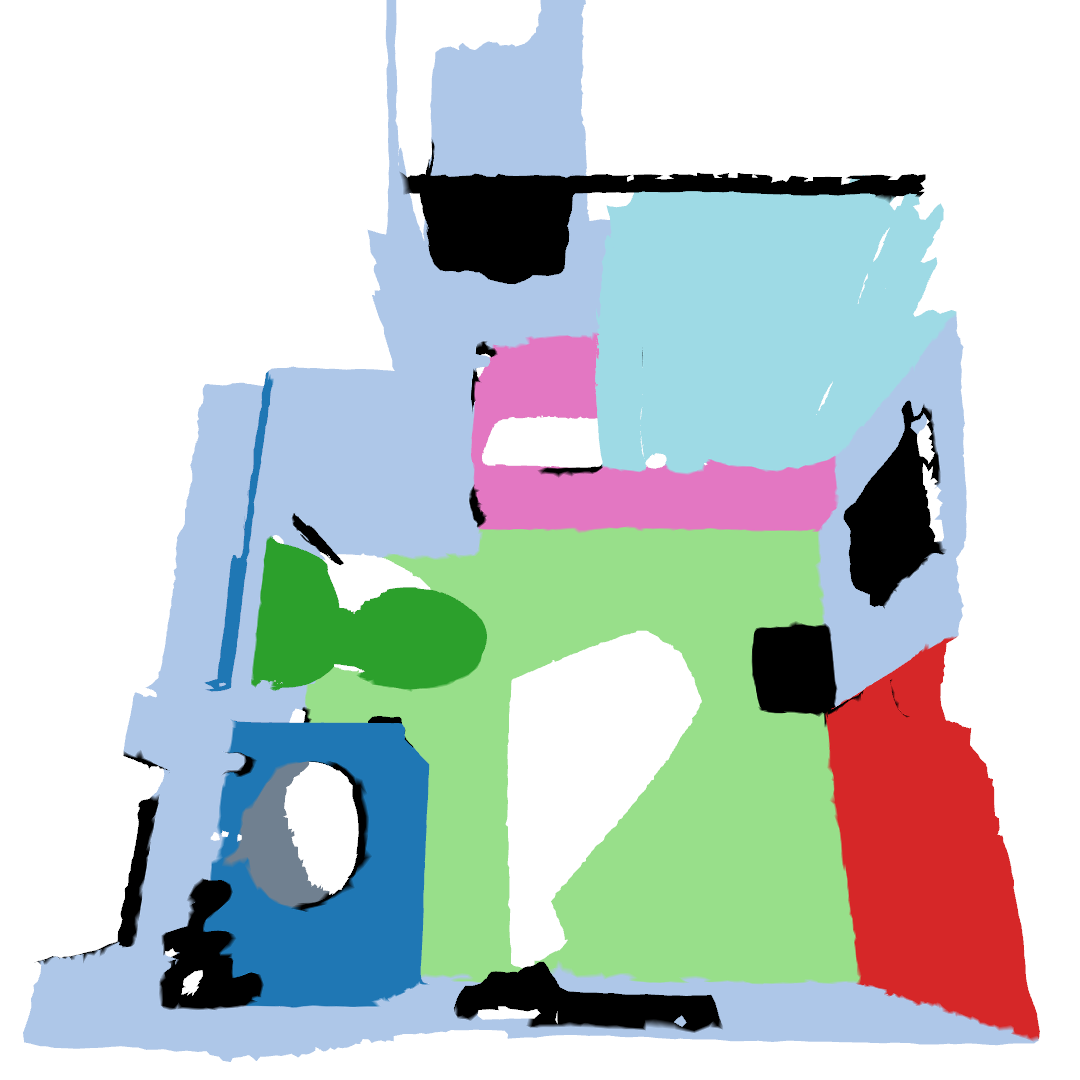}
	}
	\subfloat{
		\includegraphics[width=\len\linewidth]{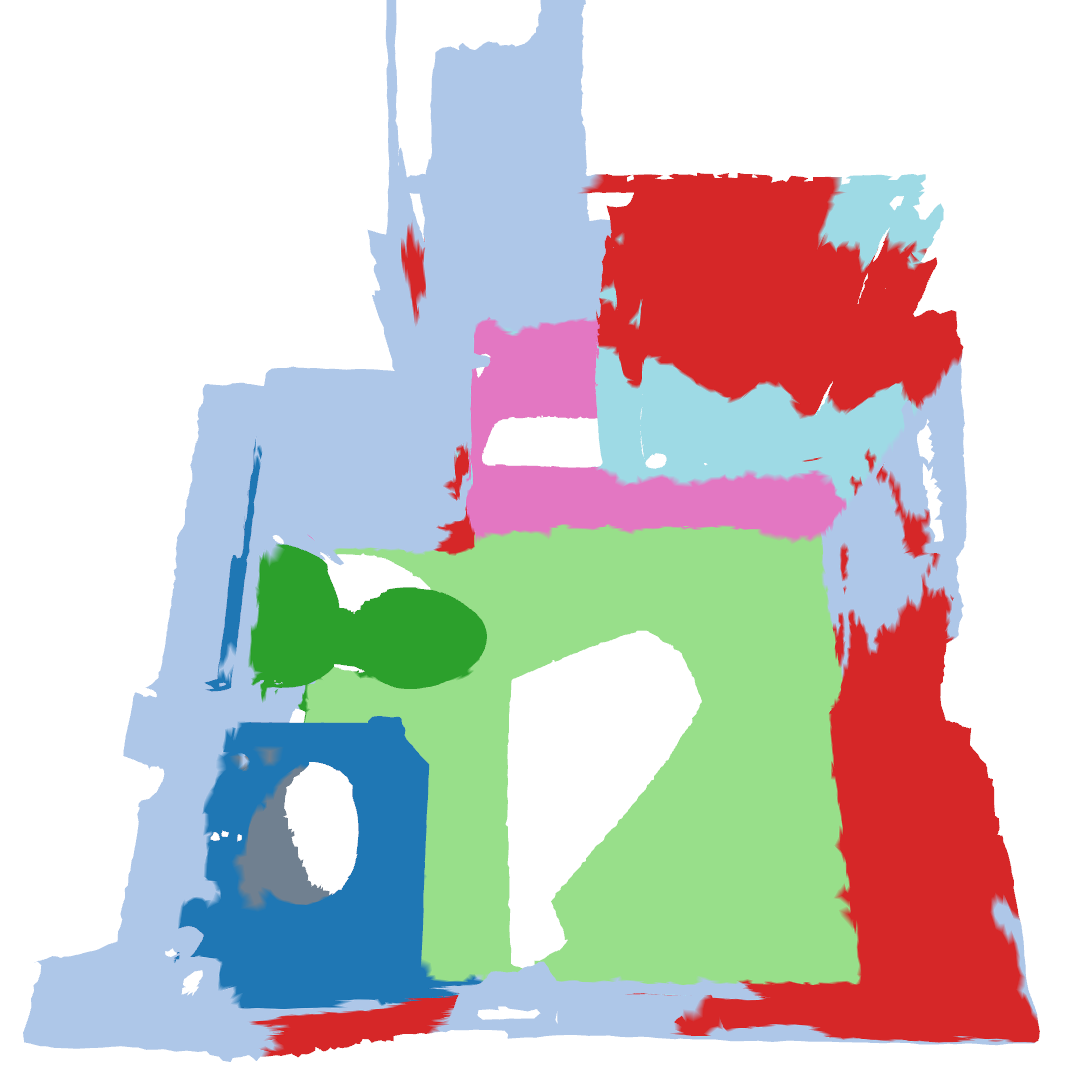}
	}
	\subfloat{
		\includegraphics[width=\len\linewidth]{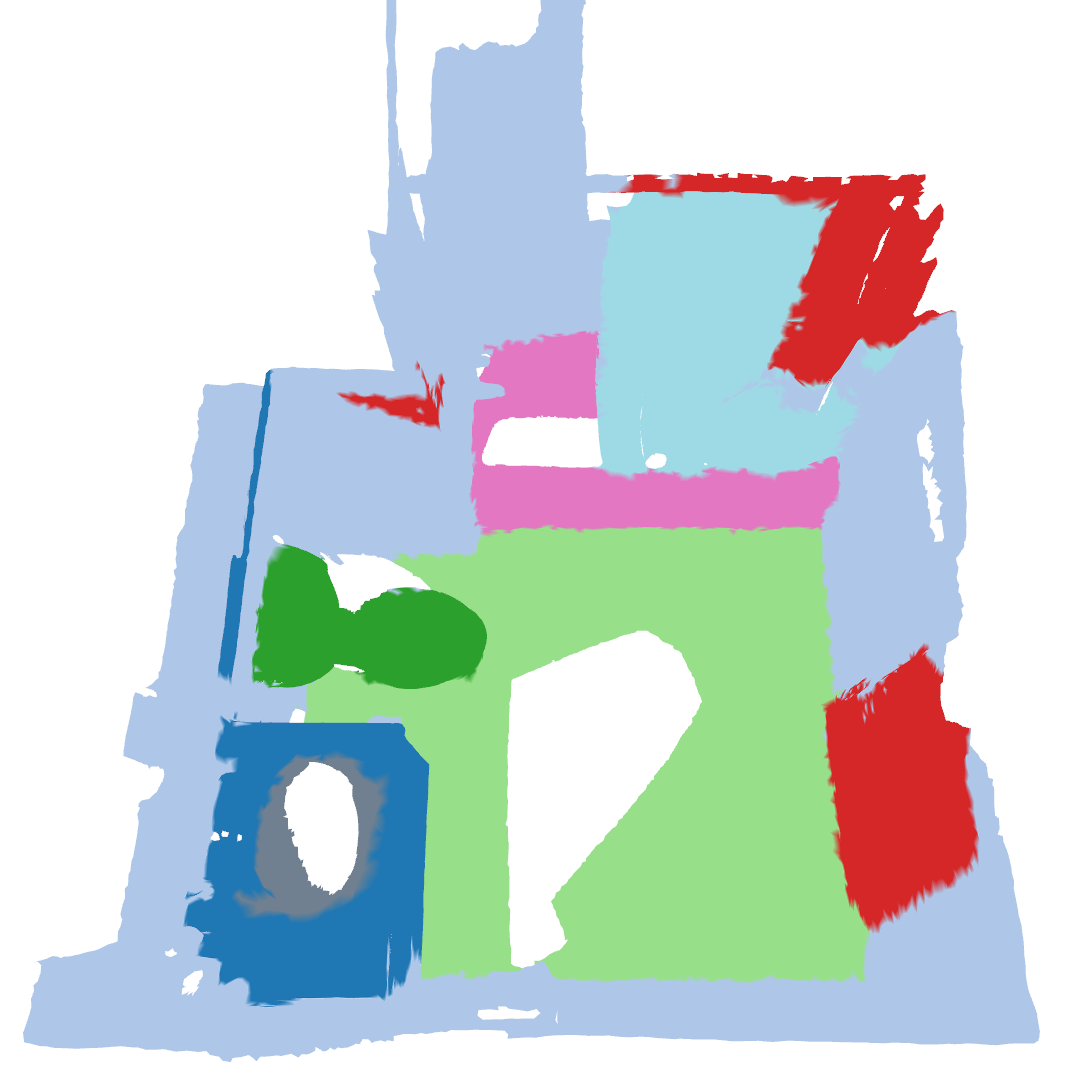}
	}
	\subfloat{
		\includegraphics[width=\len\linewidth]{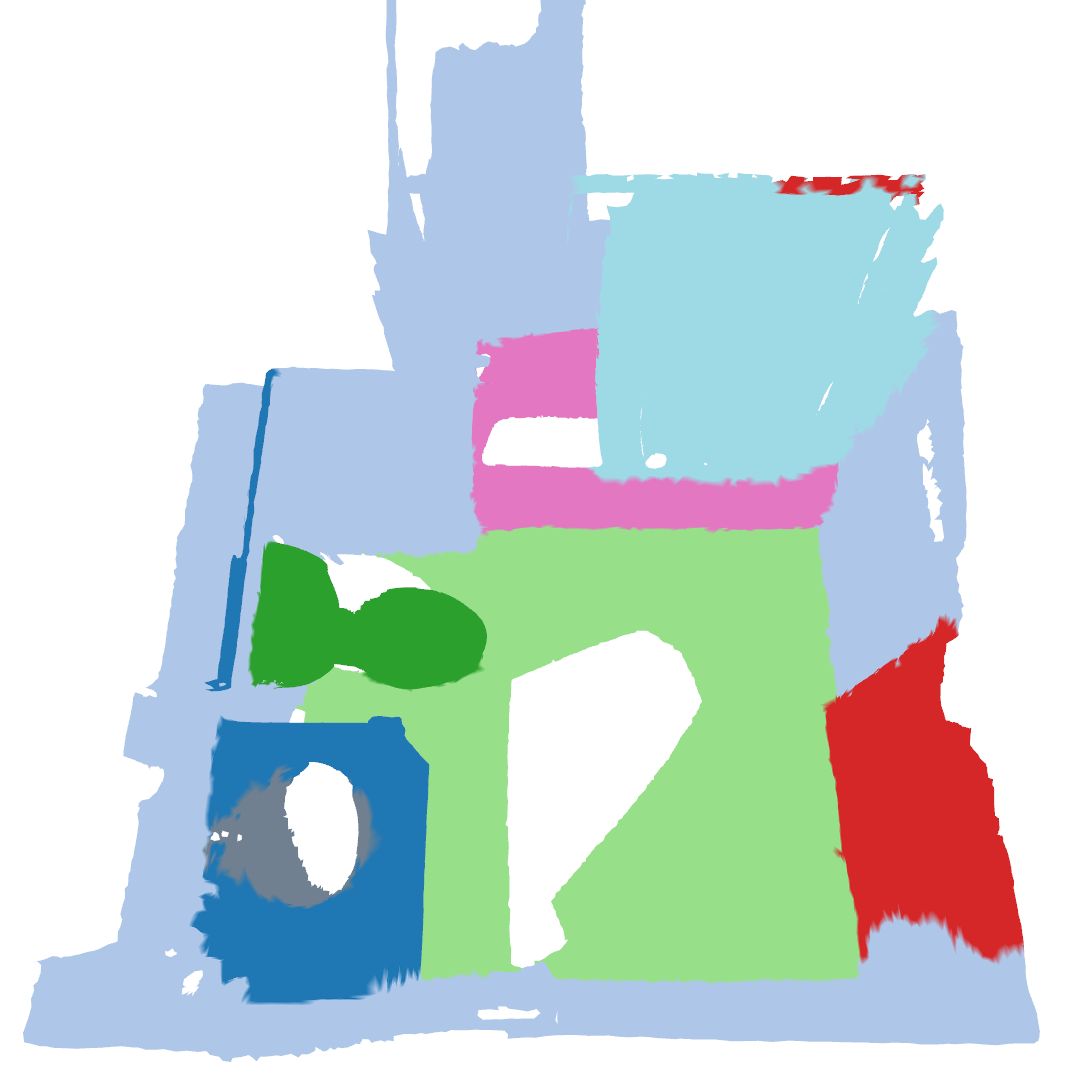}
	}

	\subfloat{
		\includegraphics[width=\len\linewidth]{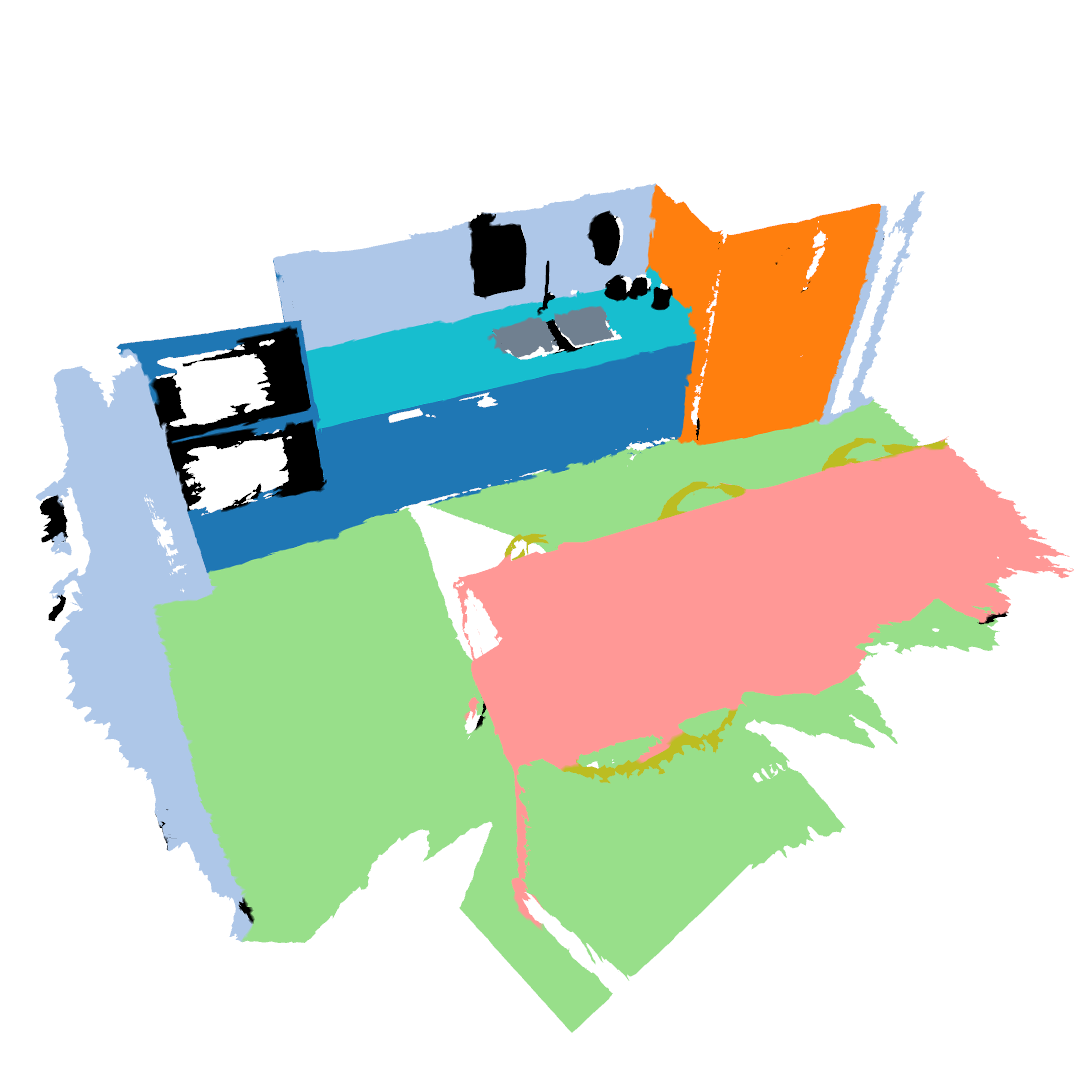}
	}
	\subfloat{
		\includegraphics[width=\len\linewidth]{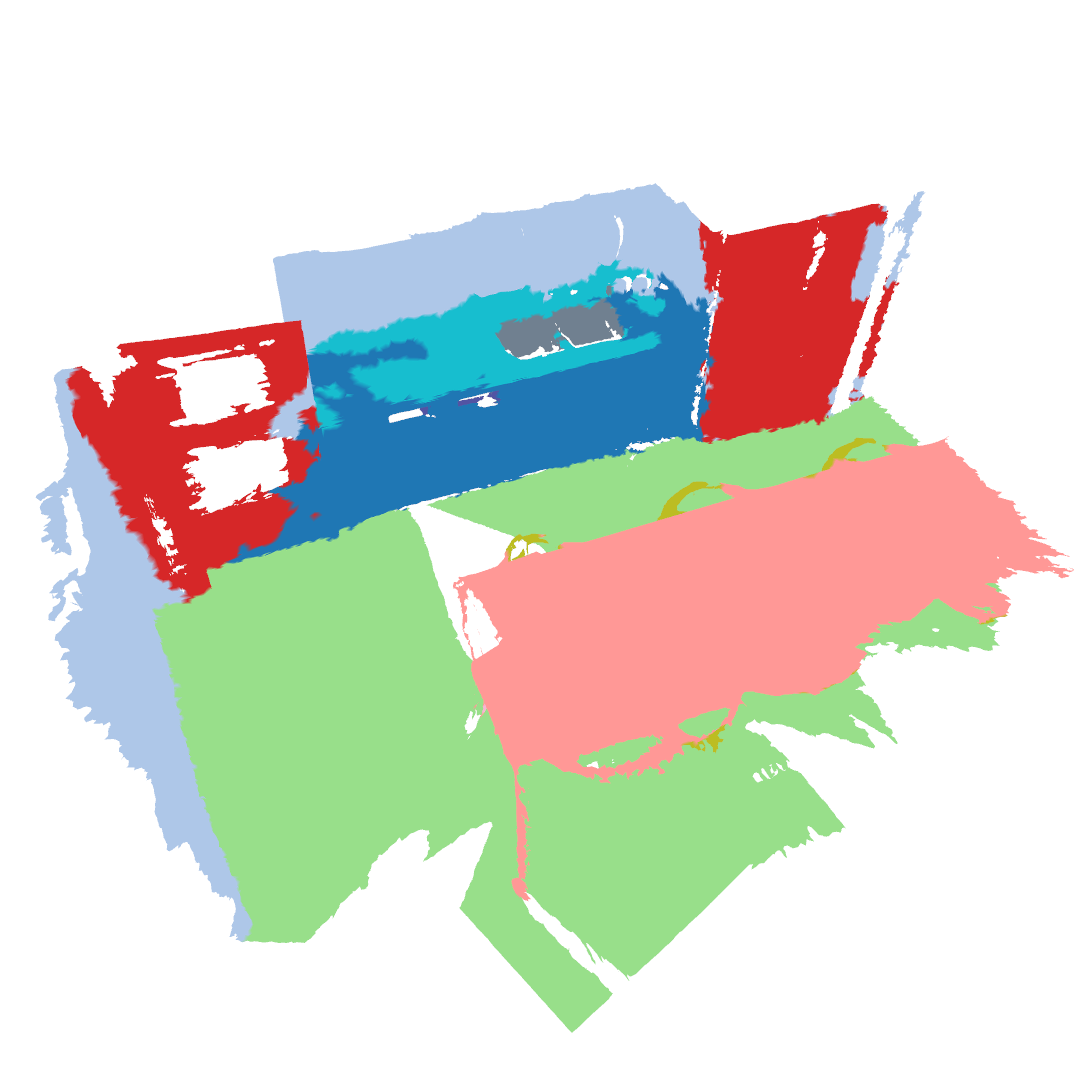}
	}
	\subfloat{
		\includegraphics[width=\len\linewidth]{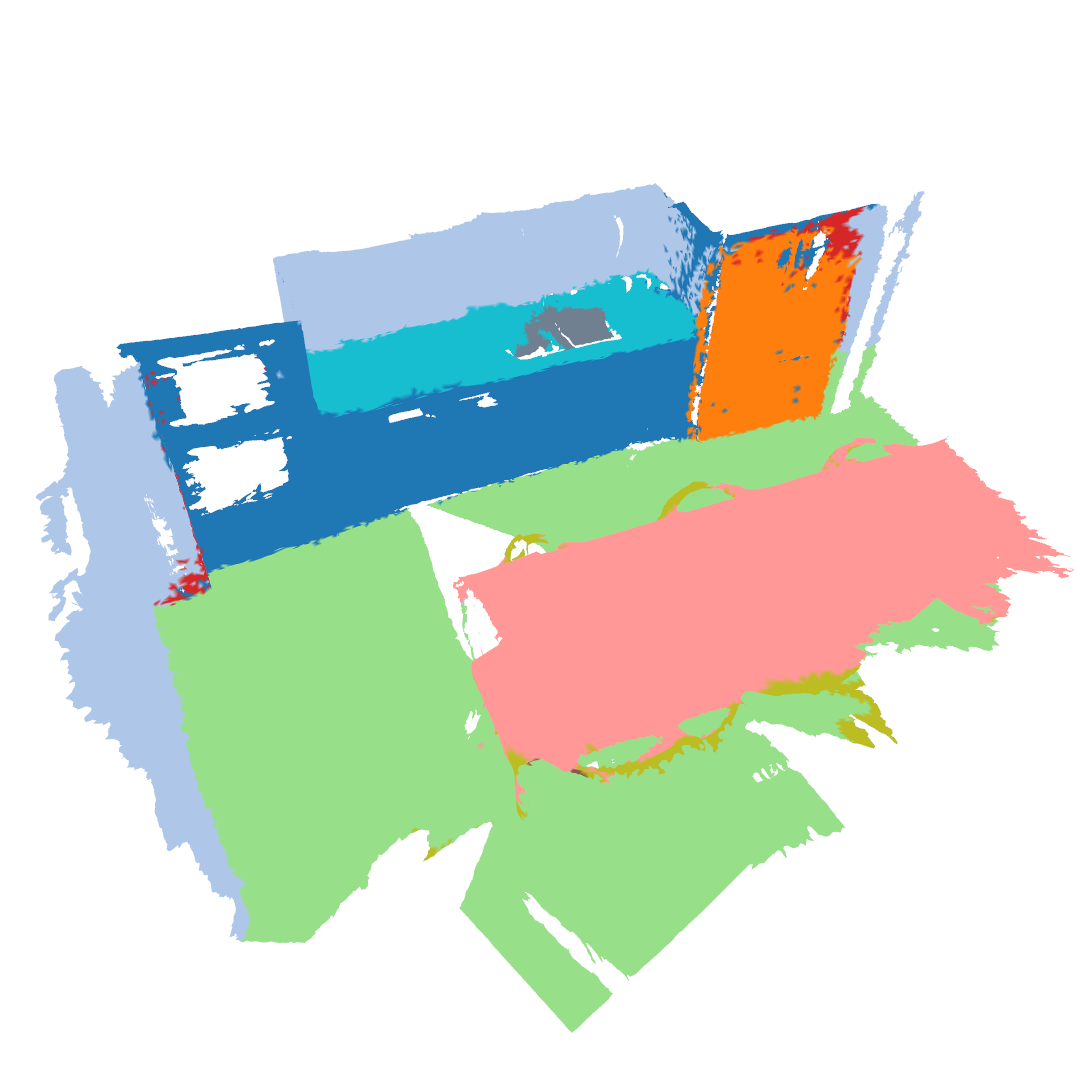}
	}
	\subfloat{
		\includegraphics[width=\len\linewidth]{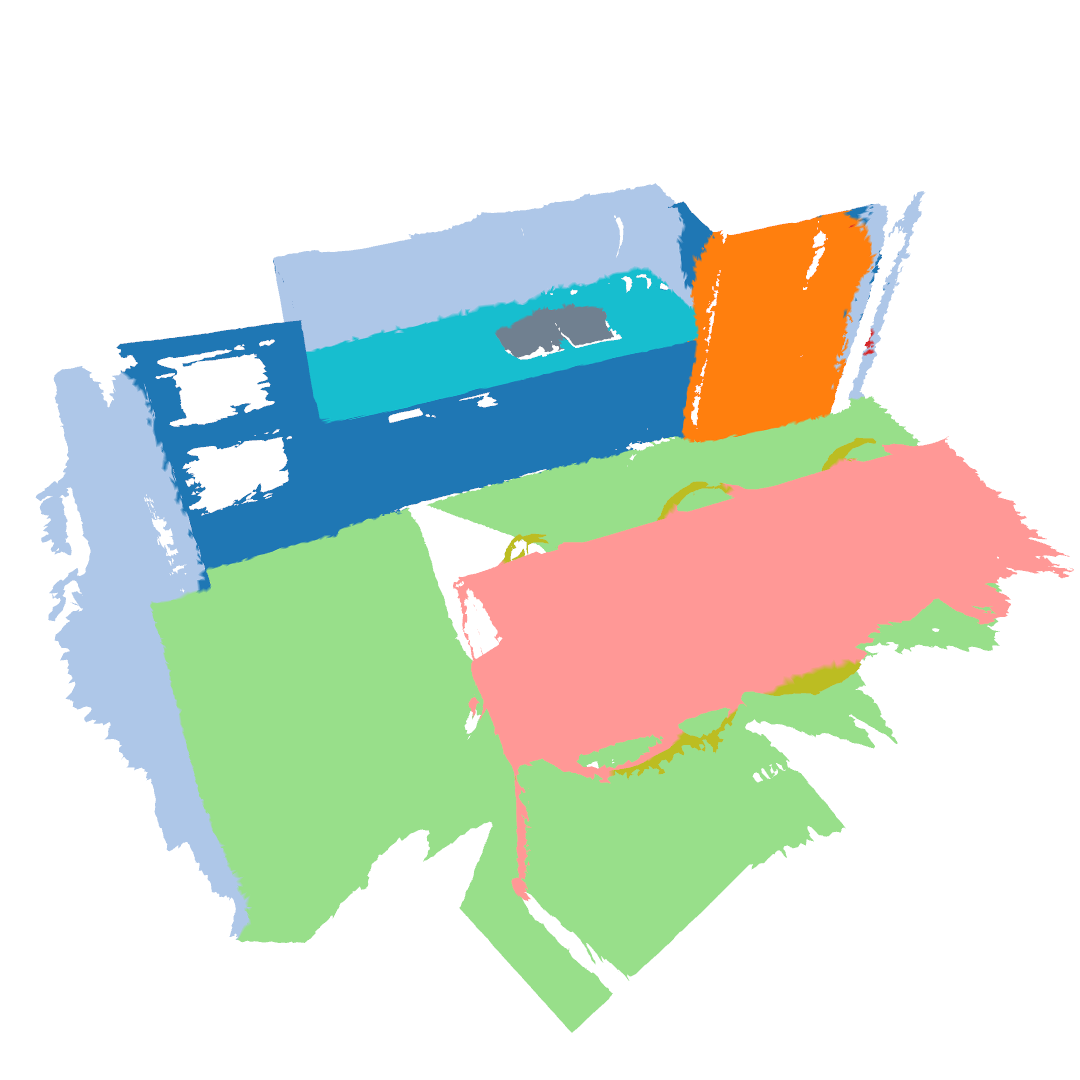}
	}

	\subfloat{
		\includegraphics[width=\len\linewidth]{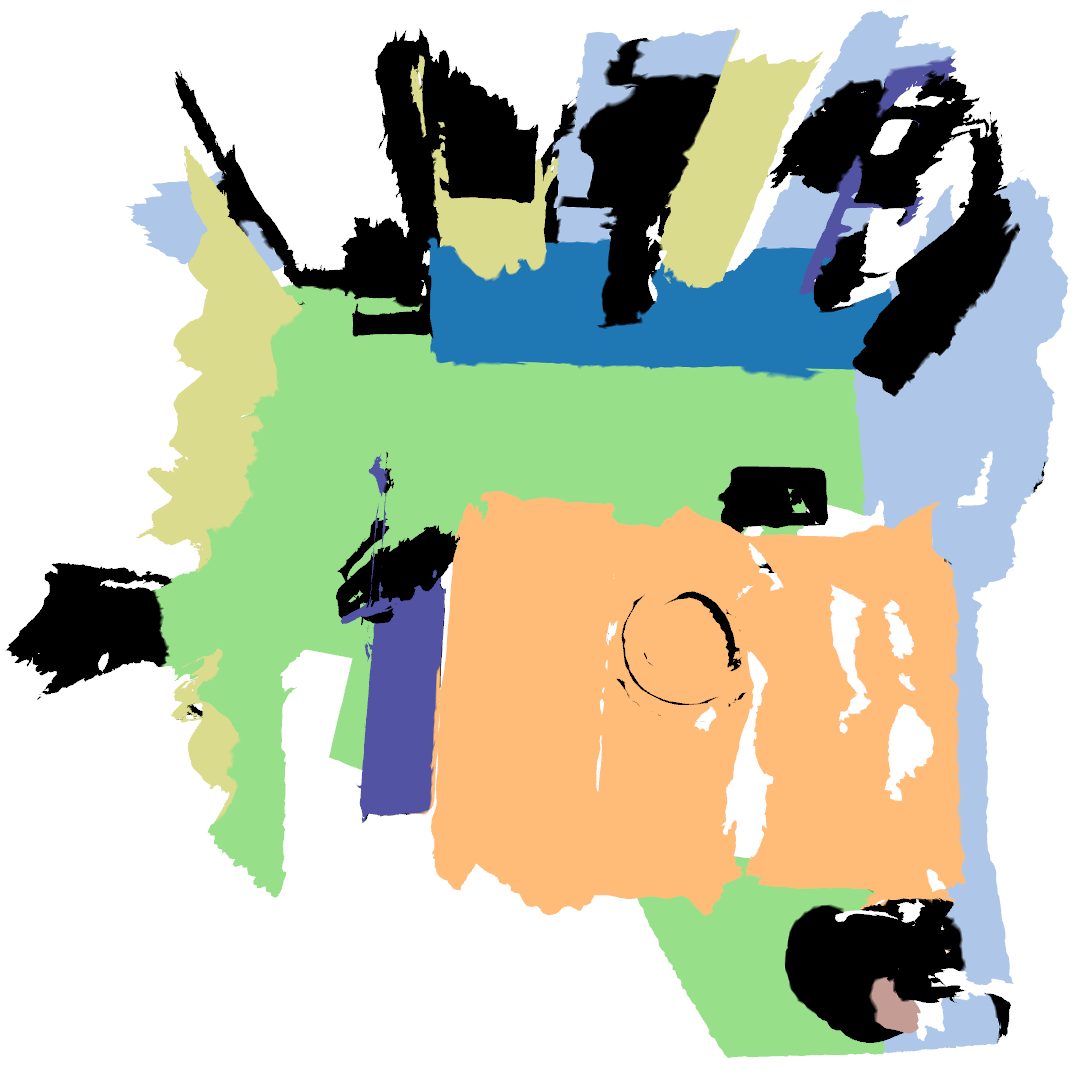}
	}
	\subfloat{
		\includegraphics[width=\len\linewidth]{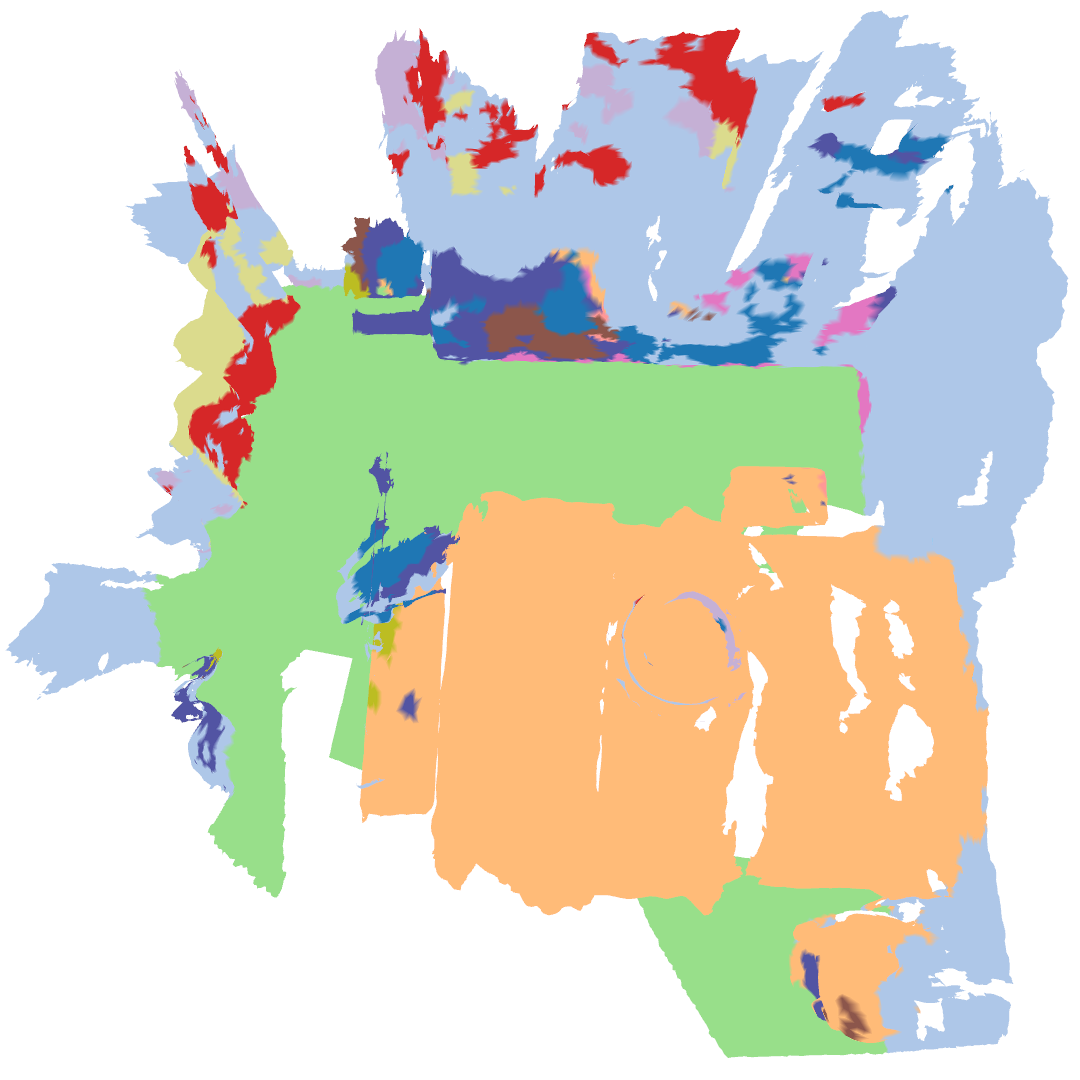}
	}
	\subfloat{
		\includegraphics[width=\len\linewidth]{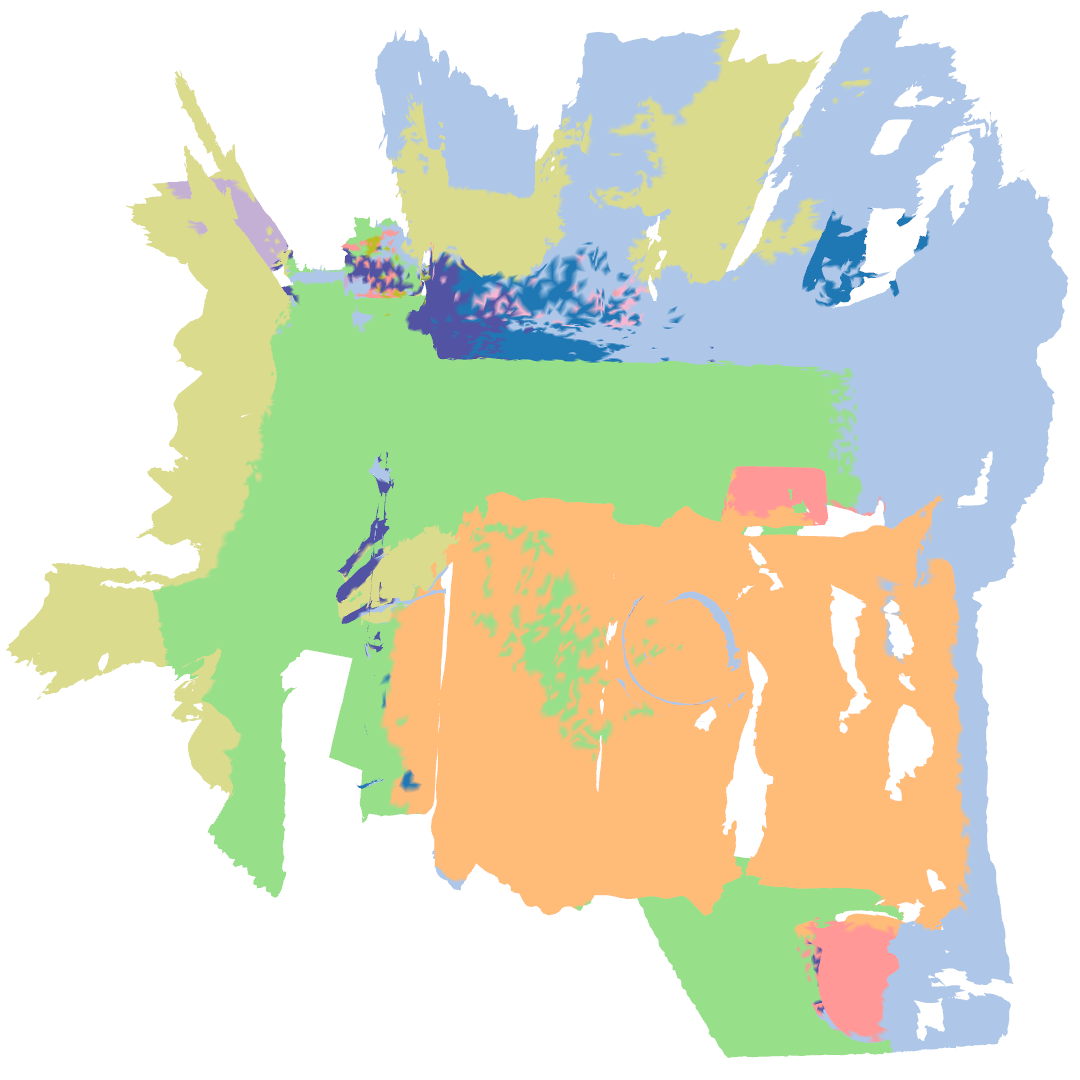}
	}
	\subfloat{
		\includegraphics[width=\len\linewidth]{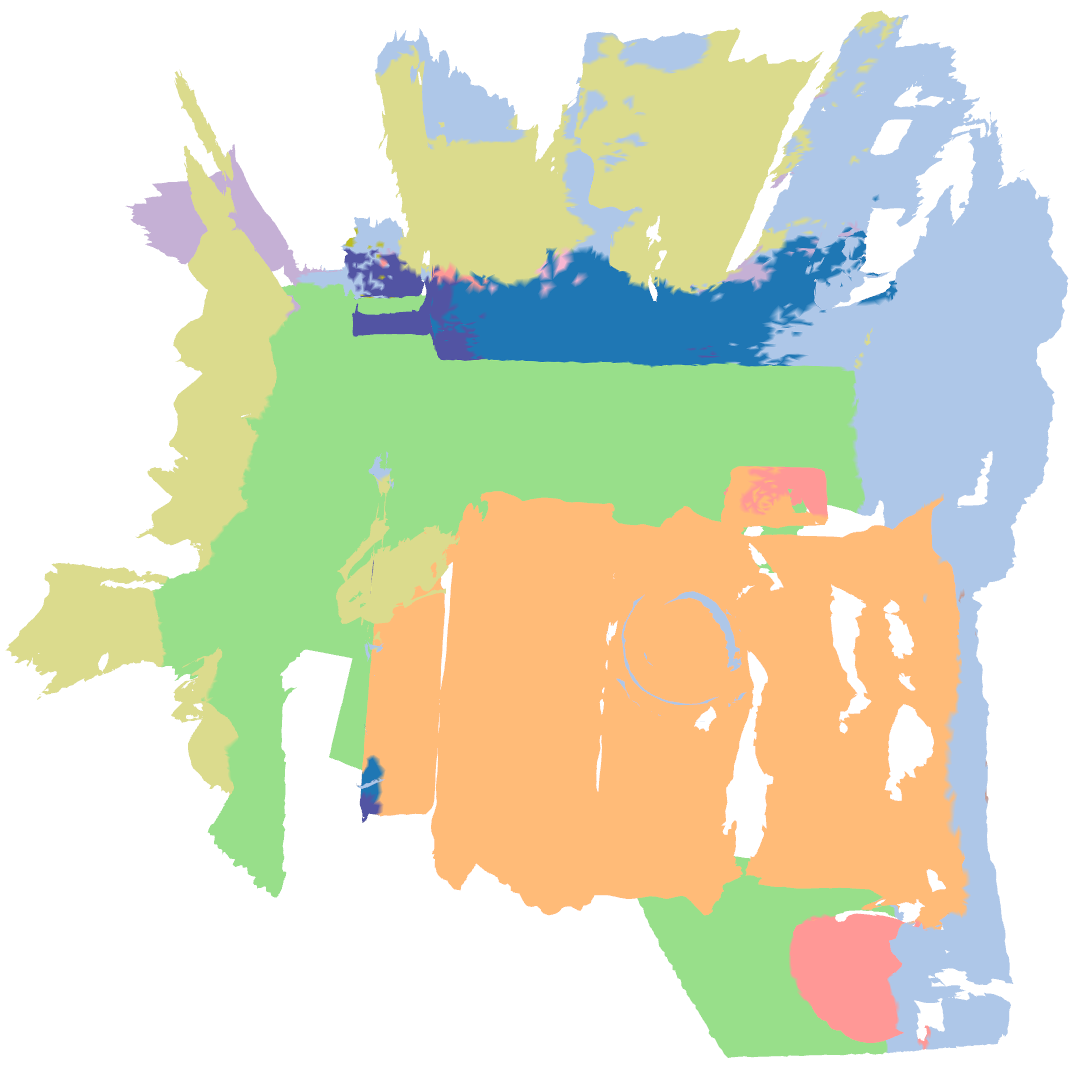}
	}

	\subfloat[Ground Truth]{
		\includegraphics[width=\len\linewidth]{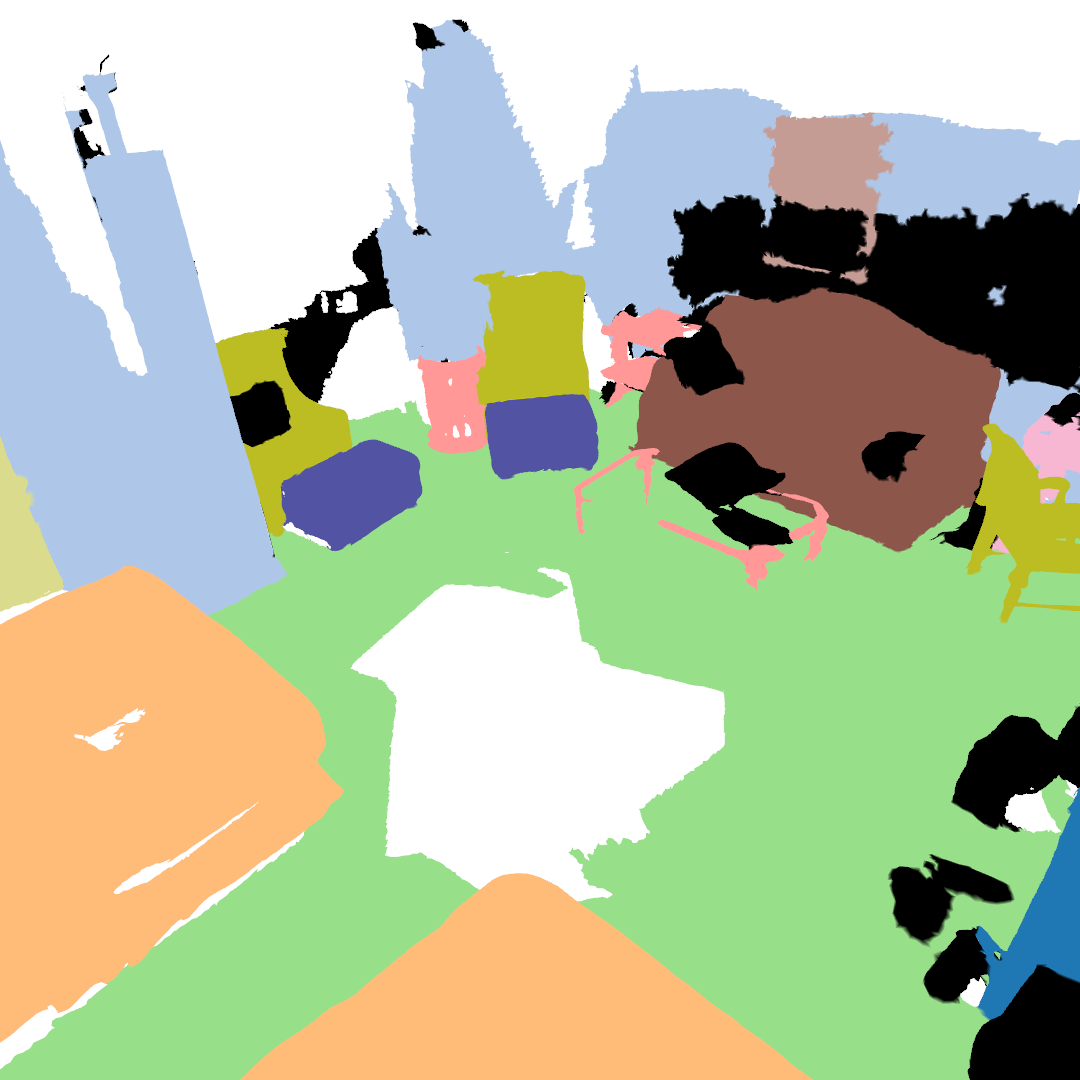}
	}
	\subfloat[PointNet++ XYZRGB]{
		\includegraphics[width=\len\linewidth]{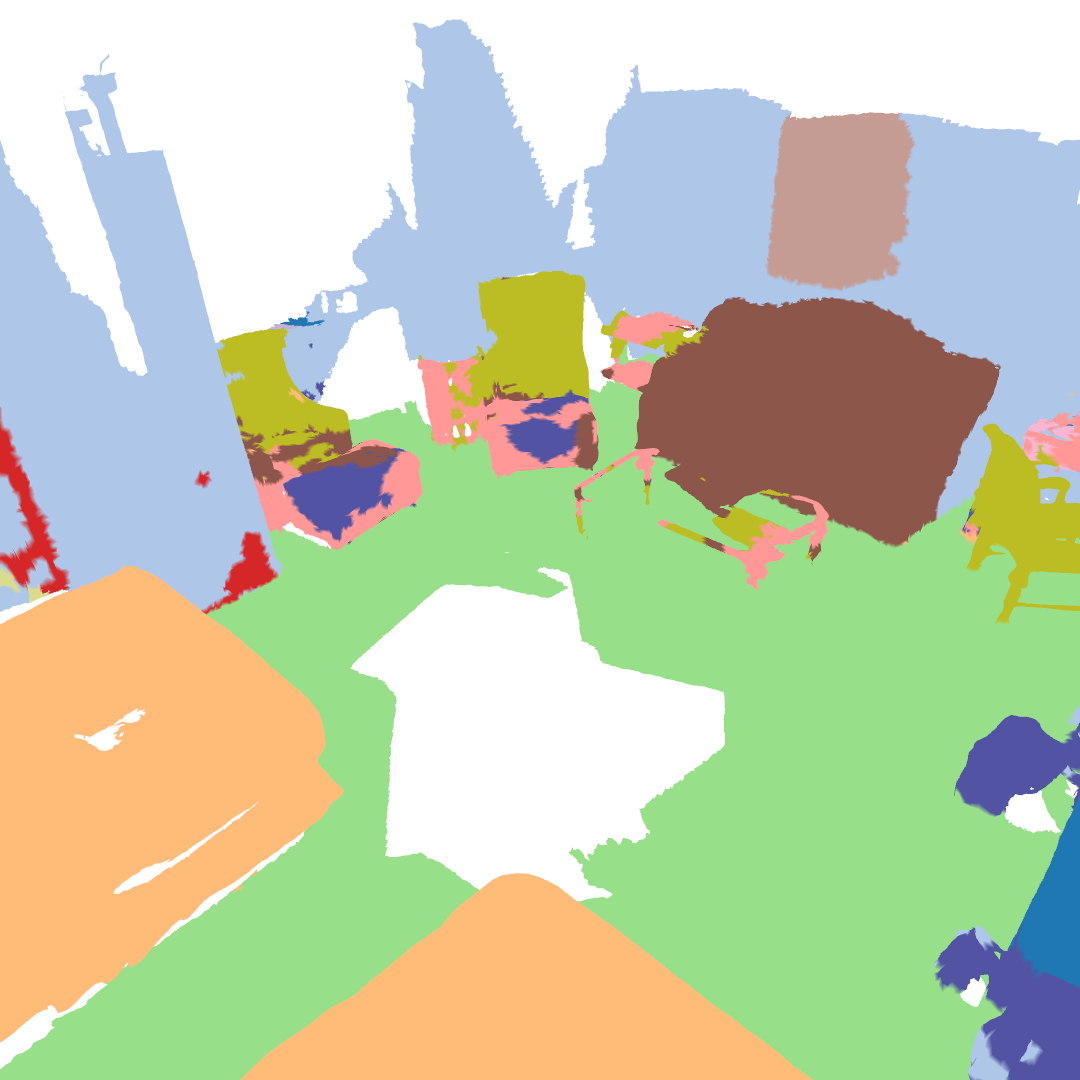}
	}
	\subfloat[Ours (2D)]{
		\includegraphics[width=\len\linewidth]{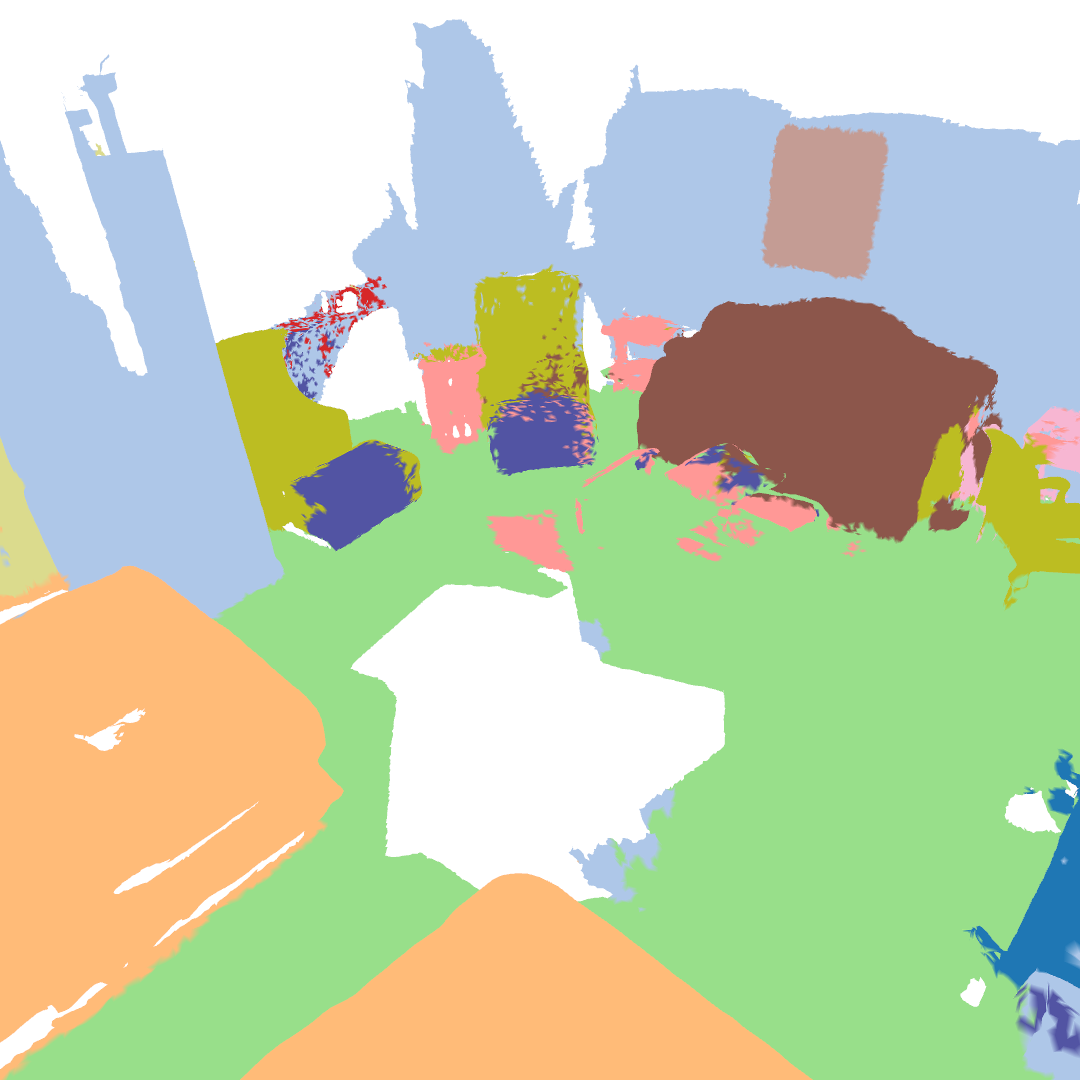}
	}
	\subfloat[Ours (2D + 3D)]{
		\includegraphics[width=\len\linewidth]{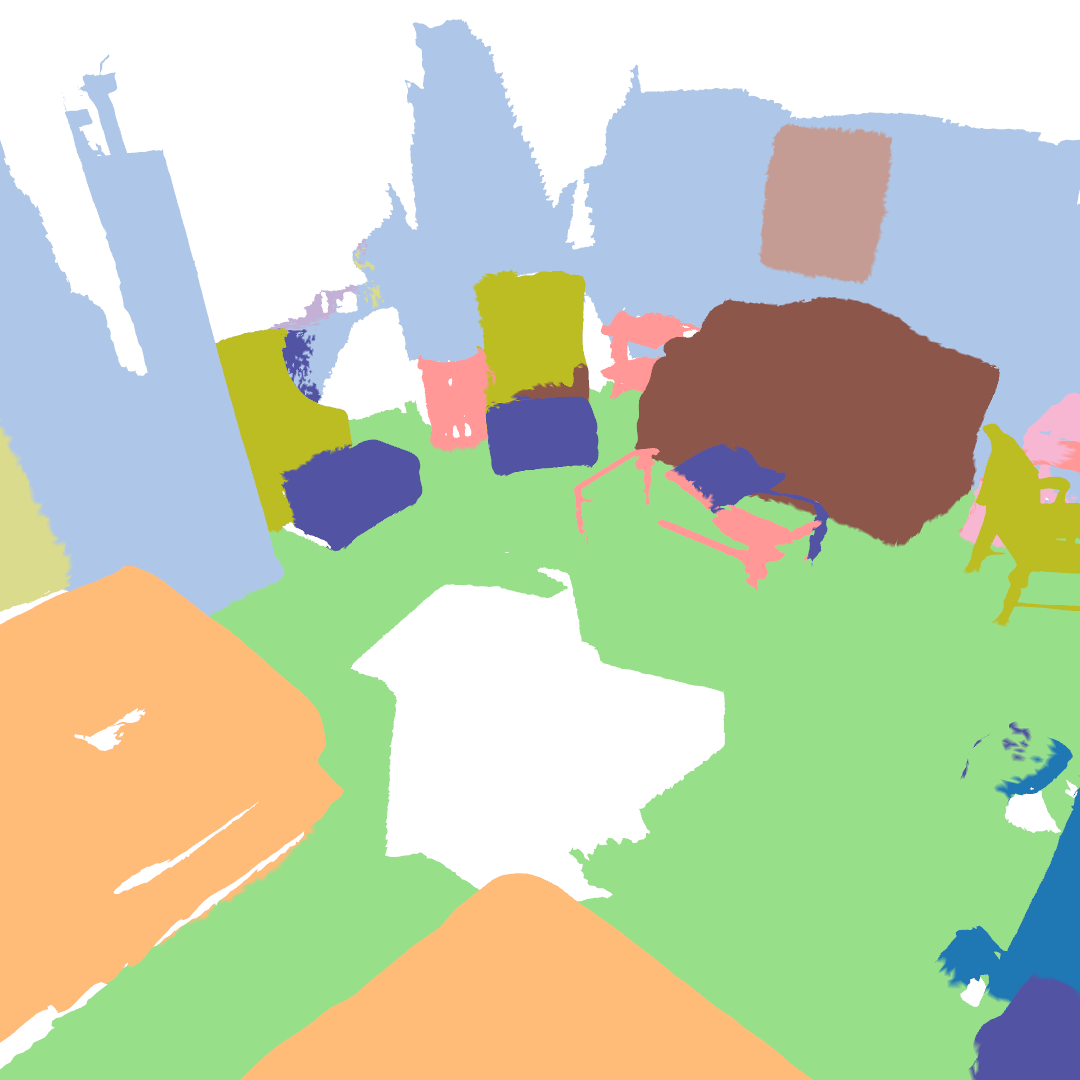}
	}
	
	\caption{Qualitative results of 3D semantic segmentation. A common error mode of PointNet++ is to misclassify similarly shaped objects (shower curtain, refrigerator, etc.) as the most prevalent door category while our method succeeds.}
	\label{fig:qualResultsSeg}
\end{figure*}

\begin{figure*}
	\centering
	\newcommand\len{0.15}
	
	\subfloat[Ground Truth]{
		\includegraphics[width=\len\linewidth]{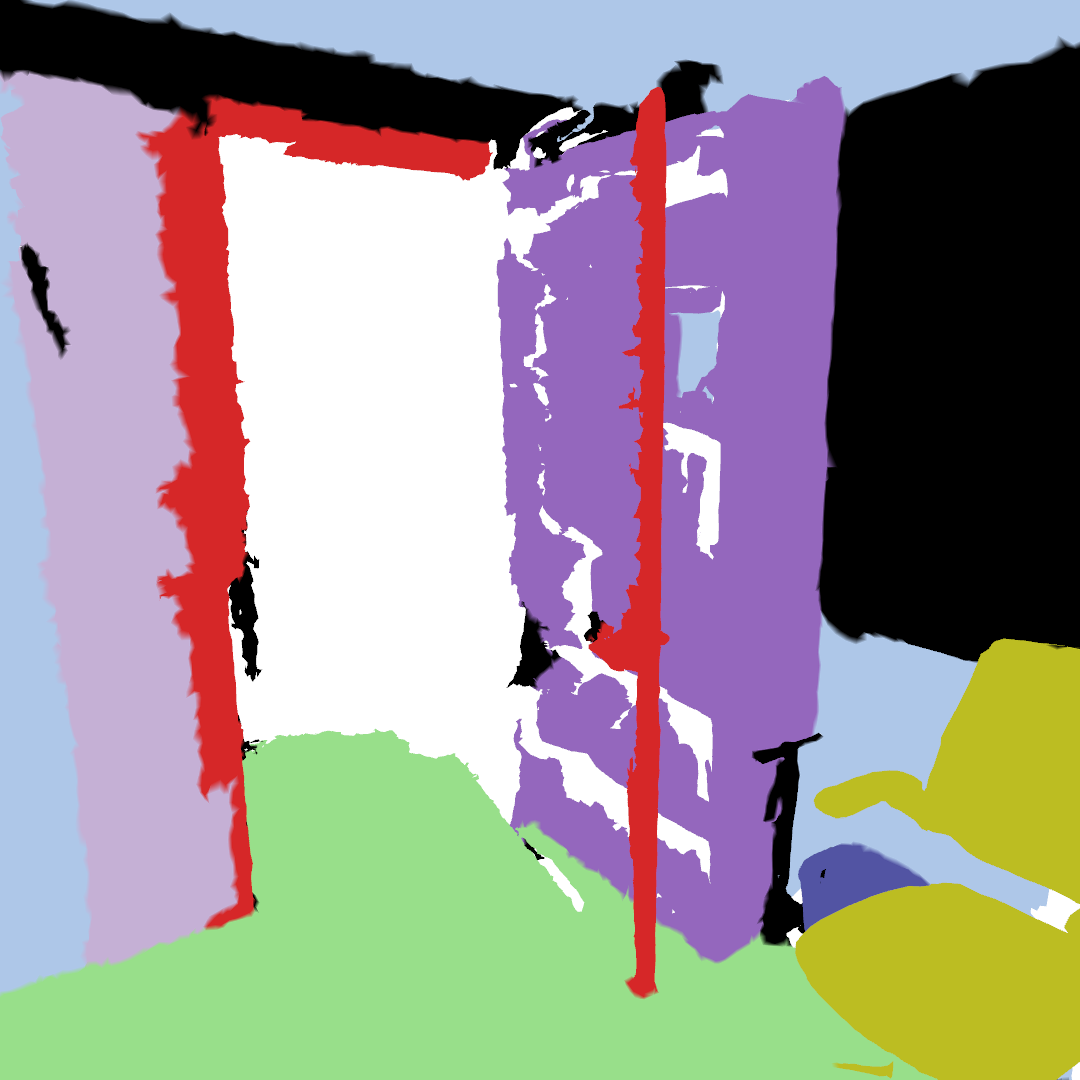}
	}
	\subfloat[PointNet++ XYZRGB]{
		\includegraphics[width=\len\linewidth]{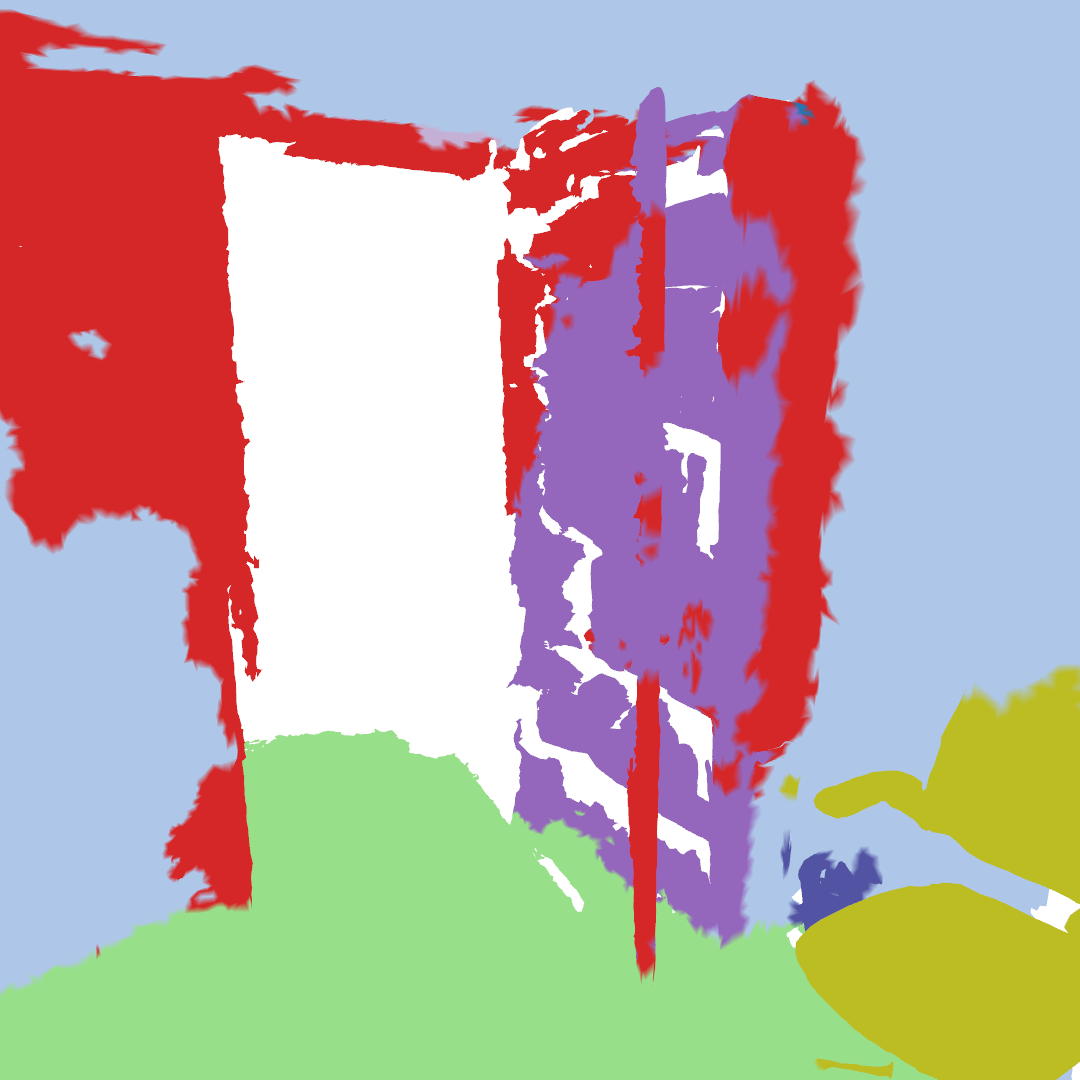}
	}
	\subfloat[Ours (2D)]{
		\includegraphics[width=\len\linewidth]{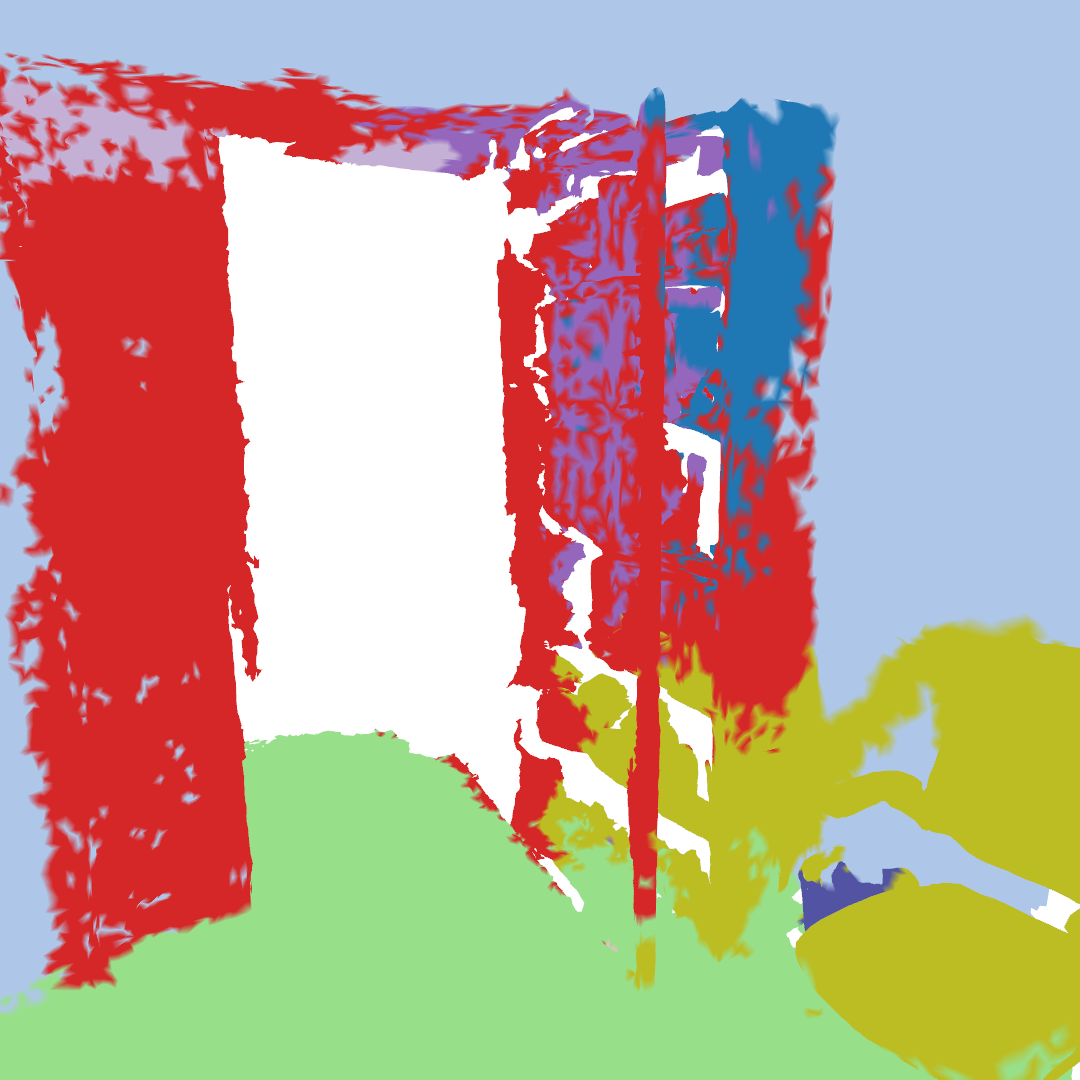}
	}
	\subfloat[Ours (2D + 3D)]{
		\includegraphics[width=\len\linewidth]{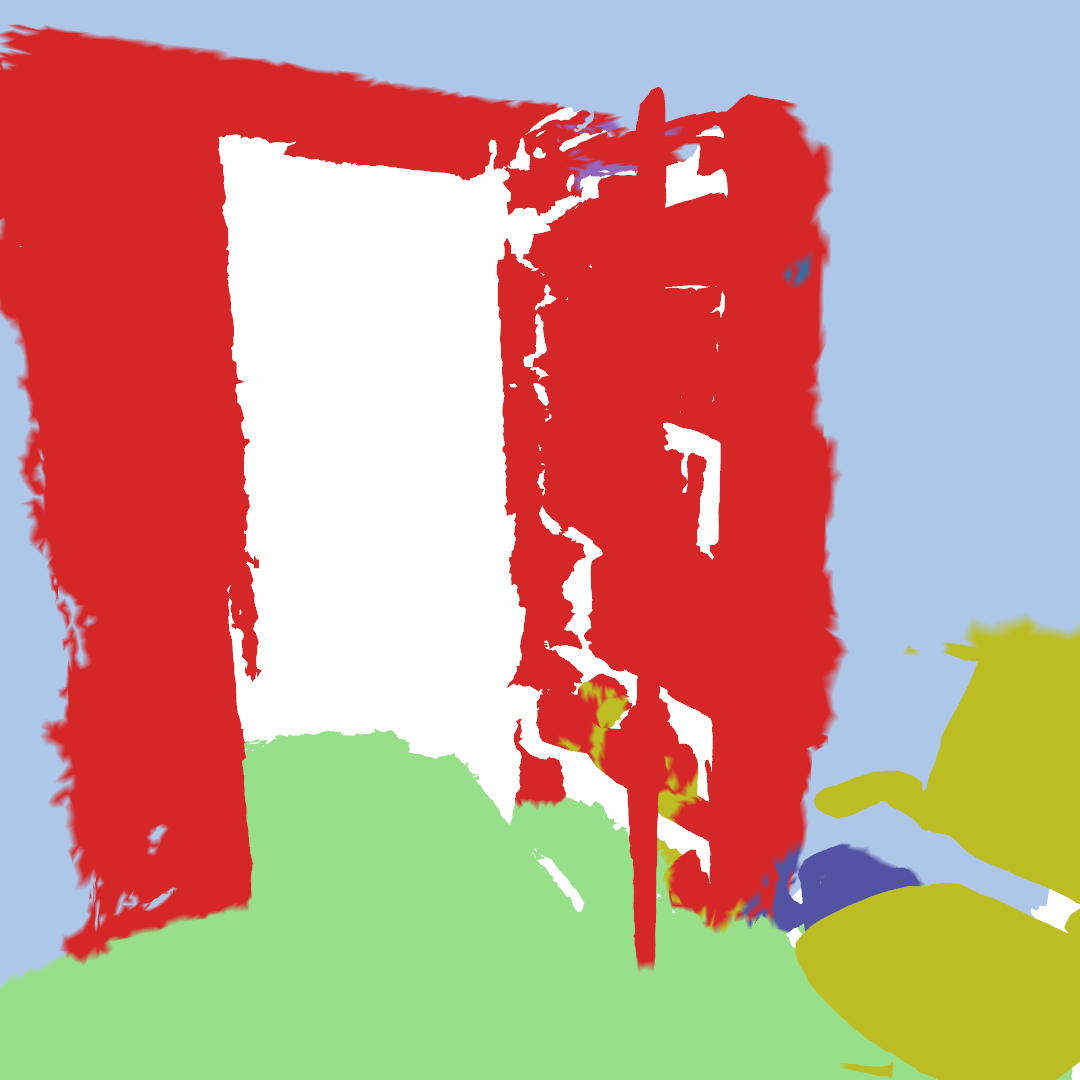}
	}
	\subfloat[color mesh (no input)]{
		\includegraphics[width=\len\linewidth]{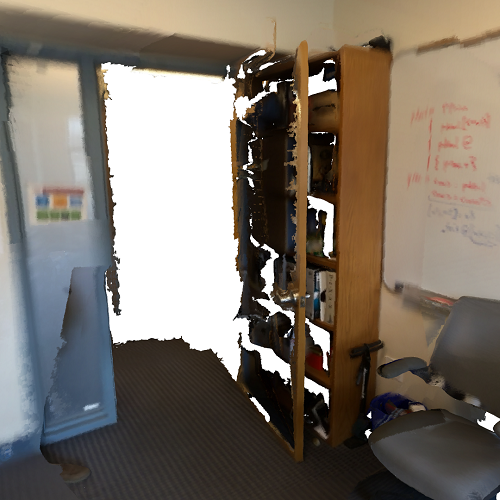}
	}
	
	\caption{Failure case of our method for 3D semantic segmentation. On the right hand side in the image is an open door next to a bookshelf. Both have very similar "wooden" appearance and are spatially close which leads our method -- which also relies on appearance information -- to misclassify the bookshelf as door.}
	\label{fig:qualResultsSegFailure}
\end{figure*}

Tab.~\ref{tab:resultsSegBenchmark-Point} shows our performance compared to the published state-of-the-art point cloud based methods on the test set.
Our MVPNet outperforms all the published point cloud based methods, like PointConv~\cite{wu2018pointconv} and PointCNN~\cite{li2018pointcnn}, by a large margin. 
This confirms the effectiveness of our approach of elevating 2D image features to 3D for geometric fusion, especially for classes with flat shapes, i.e. refrigerator, picture, curtain and the like, which lack discriminative geometric cues for point cloud based networks.
Qualitative results are shown in Fig.~\ref{fig:qualResultsSeg} and a failure case in Fig.~\ref{fig:qualResultsSegFailure}.

Tab.~\ref{tab:resultsSegBenchmark-Voxel} shows our performance compared to the published state-of-the-art voxel based methods on the test set. 
3DMV~\cite{dai20183dmv} is a joint 2D-3D network similar to ours, but in the voxel-based domain, and does not match our performance and inference time (500 s/scene vs. 3.35 s/scene).

Although there exists a gap between our result and SCN~\cite{graham20183d}, our MVPNet is much more robust to low resolution point clouds as detailed in Sec.~\ref{sec:robustness}. This is relevant to robotic applications where, due to sensor limitations, point clouds are always much sparser than images. Furthermore, we compare with SCN in terms training/inference time and number of parameters in Tab.~\ref{tab:scn}. It shows that MVPNet is comparable with highly engineered SCN. However, MVPNet is able to converge in 20 hours on a GTX 1080Ti (incl. pretrain of 2D encoder-decoder), while it takes 12 days for heavyweight SCN for the same GPU model, or 4 days even with a V100.
\begin{table}
	\scriptsize
	\centering
	\setlength{\tabcolsep}{0.25em}
	\begin{threeparttable}
	\begin{tabular}{lccccc}
		\toprule
		Method & mIoU & batch size & train time & forward time/scene\tnote{*} & \#parameters\\
		\midrule
		SCN (light) & 57.5 & 32 & 18h & 0.194s & 2.7M \\
		SCN (heavy) & 68.2 & 4 & $>$12d & 2.21s & 30.1M \\
		\midrule
		ResNet (2D) & - & 32 & 8h & - & 23M \\
		MVPNet (3-view) & 65.9 & 32 & 12h & 2.22s & 0.98M \\
		MVPNet (5-view) & 67.3 & 32 & 18h & 3.35s & 0.98M \\
		\bottomrule
	\end{tabular}
\begin{tablenotes}
\item[*] Preprocessing time not included.
\end{tablenotes}
\end{threeparttable}
\caption{Runtime comparison with SCN on a GTX 1080Ti.}
\label{tab:scn}
\end{table}

\subsection{Robustness to Varying Point Cloud Density}\label{sec:robustness}
\begin{figure}
	\centering
	\includegraphics[width=0.8\linewidth]{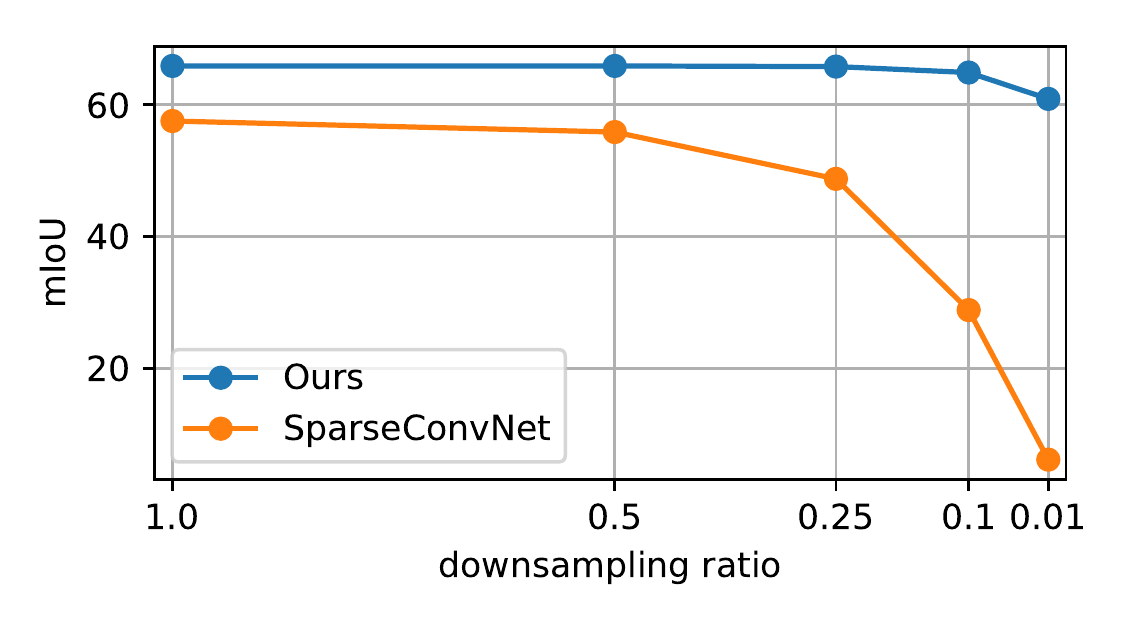}
	\caption{Robustness to input point cloud density of our method compared to SparseConvNet~\cite{graham20183d}. The x-axis shows the ratio of points that are kept and the y-axis shows the mIoU on the validation set of ScanNetV2.}
	\label{fig:robustness}
\end{figure}

Real-world 3D sensors such as lidars and depth cameras have much lower resolution than 2D RGB cameras and the point cloud density also varies with view point angle, lighting conditions, object-to-sensor distance and object surface reflectivity. Thus, it is important for algorithms to be robust to varying point cloud density at test time, because training data can hardly cover all cases.
To examine robustness to sparsity, we uniformly subsample the whole scene point cloud and feed it to the networks that were trained on full resolution. We report the results in Fig.~\ref{fig:robustness}.
While our MVPNet is hardly affected at lower resolutions, the performance of SparseConvNet (SCN)~\cite{graham20183d} suffers severely.

We attribute the performance difference mainly to two factors.
First, the quality of our image features is not deteriorated when the point cloud is downsampled, because they are computed in the original dense 2D image. During 2D-3D lifting, even if few unprojected image pixels are finally used at coarse point cloud resolution, the image features can maintain their quality thanks to the receptive field of the 2D encoder-decoder. This is not the case for SCN where only the sparse RGB information at each 3D point is used.
The second factor is related to the different neighborhood definition in voxel grids and point clouds. Voxel-based methods such as SCN have fixed neighbors defined by the discrete grid. Point-based methods on the other hand, use continuous locations and in each network layer neighbors are sampled, e.g. with ball query, that adapt naturally to the local point distribution. This enables our PointNet++ based approach to cope well with varying point cloud density.

\subsection{Extension to 3D Instance Segmentation}\label{sec:3dInstSeg}
In order to assess the ability of our MVPNet to be applied to a different task, we extend it to 3D Instance Segmentation on ScanNetV2. We use trained model of MVPNet for semantic segmentation to predict all scenes of the train/val/test set and save the features of the last layer before the segmentation head to disk. We modified R-PointNet~\cite{yi2018gspn} in order to take the semantic features from MVPNet as input and yield a significant improvement from 38.8 to 47.1 mAP on the validation set which demonstrates the versatility of MVPNet.

\section{Ablation Studies}
To analyze our design choices and provide more insights, we conduct ablation studies on the validation set of ScanNetV2 and report the average IoU (mIoU) over 20 classes. 
The 2D encoder-decoder network is frozen in order to accelerate training, since we observe no significant improvement with end-to-end training.
Unless stated otherwise, our 2D backbone is VGG16~\cite{simonyan2014very} and our 3D backbone contains 4 set abstraction and 4 feature propagation layers. The numbers of centroids are 1024, 256, 64, 16 respectively.

\subsection{Number of Views}
\begin{table}
	\scriptsize
	\centering
	\begin{tabular}{lccccc}  
		\toprule
		Number of frames & 1 & 3 & 5\\
		\midrule
		Average coverage & 68.1 & 92.9 & 97.4\\
		\midrule
		mIoU & 62.8 & 64.5 & 64.8\\
		\bottomrule
	\end{tabular}
	\caption{Average coverage and mIoU as a function of the number of unprojected views.
	Results on validation set of ScanNetV2.}
	\label{tab:coverage}
\end{table}
We define \textit{coverage} as the ratio of points in the input point cloud that have at least one unprojected neighbor point with image features at a distance less then 0.1m.
Tab.~\ref{tab:coverage} shows how the number of views affects coverage and mIoU.
We removed the feature aggregation module for this experiment.
With 1 view the coverage reaches 68.1\%, and with 3 frames already exceeds 90\%.
More views lead to higher mIoU, but introduce more computation. We choose 3 frames as default in this trade-off.

\subsection{Feature Aggregation Module}
\begin{table}
	\scriptsize
	\centering
	\begin{tabular}{ccccc}
		\toprule
		Number of views & k-nn & MLP & Aggregation & mIoU \\
		\midrule
		1 & 1 & w/o & none & 61.7 \\
		1 & 3 & w/o & sum & 62.8 \\
		1 & 3 & w/ & sum & 62.5 \\
		\midrule
		3 & 1 & w/o & none & 64.5 \\
		3 & 3 & w/o & sum & 64.5 \\
		3 & 3 & w/ & sum & \textbf{65.0} \\
		3 & 3 & w/ & max & 64.7 \\
		\bottomrule
	\end{tabular}
	\caption{Effect of feature aggregation. Results on validation set of ScanNetV2.}
	\label{tab:featAggregation}
\end{table}
In the following we study the parameters of our Feature Aggregation Module, defined in eq.~\ref{eq:featAggreg}, which distills features from the unprojected point cloud $\mathcal{S}_{\text{dense}}$, obtained from multiple views, into the input point cloud $\mathcal{S}_{\text{sparse}}$.
We report our results in Tab.~\ref{tab:featAggregation} for 1 and 3 views, 1 or 3 nearest neighbors feature sampling, with or without MLP, and we also try maximum instead of sum as feature pooling function.

For the 1-view case, using 3 nearest neighbors instead of only 1 increases the performance by at least 0.8 mIoU. 
Due to the limited coverage of a single view, far-away image features are sampled for uncovered points and multiple neighbors might alleviate this problem by analyzing feature consistency between them.

For the 3-view case, we find that the number of nearest neighbors does not affect performance. This might be because coverage is already very high (92.9\%) which means that, as opposed to the 1-view case, features can always be sampled from close-by. We also try summation instead of maximum to pool the features which does not significantly change the results. Using an MLP can slightly improve, maybe because it can transform 2D image features to an embedding space more consistent with the 3D representation. Our final choice for all other experiments is 3 nearest neighbors with MLP and sum aggregation.

\subsection{Fusion}

\begin{table}
	\scriptsize
	\centering
	\begin{tabular}{lc}
		\toprule
		Method & mIoU\\
		\midrule
		PointNet++ (XYZ) [baseline] & 54.5\\
		PointNet++ (XYZRGB) [baseline] & 57.8\\
		\midrule
		2D CNN & 57.2\\
		Ours (late fusion) & 58.4\\
		Ours (intermediate fusion) & 64.8\\
	    Ours (early fusion) & \textbf{65.0}\\
	    \midrule
	    Ours (w/o xyz) & 62.8\\
		\bottomrule
	\end{tabular}
	\caption{Effect of multiple modalities and different strategies of fusion. Results on validation set of ScanNetV2.}
	\label{tab:ablationFusion}
\end{table}
In this section we want to answer the question how to best fuse geometry and image features with point cloud based networks and give an insight about the strength of each modality. In Tab.~\ref{tab:ablationFusion} we report our quantitative results on the validation set. 
Our PointNet++ baseline yields 54.5 mIoU with XYZ only and 57.8 mIoU with additional color information.

In order to assess the strength of multi-view vs. geometry features, we conduct an experiment with 3 views where we unproject the output semantic labels of the pretrained 2D encoder-decoder to 3D and attribute the nearest neighbor 2D label to each 3D point in the input point cloud.
This multi-view 2D CNN approach can already achieve similar performance as PointNet++ on colored point clouds, which confirms the benefit of features computed on dense 2D images before 2D-3D lifting.

Next, we study three fusion strategies introduced in Sec.~\ref{sec:3dFusion}.
We could yield slightly better performance with the late fusion approach (+1.2 mIoU) than with the 2D CNN baseline.
Intermediate fusion leads to much better results (+7.6 mIoU) than late fusion.
Early fusion can reach the best score (+7.8 mIoU), and uses less parameters and computation compared to intermediate fusion.
The observation is different from the voxel-based method 3DMV\cite{dai20183dmv}, where geometric features and image features are concatenated late, at roughly 2/3 in the network.

Moreover, we investigate whether it is necessary to add geometric features (XYZ coordinates) in the early fusion approach or if the image features are sufficient. In fact, PointNet++ already induces a geometric hierarchy and the dimension of the XYZ coordinates is much smaller compared to that of the image features (3 vs. 64).
Nonetheless, the obtained result (-2.2 mIoU without XYZ) proves the contrary and indicates that MVPNet actually benefits from geometric features as complementary information to images.

\subsection{Stronger backbone}
\begin{table}
	\scriptsize
	\centering
	\begin{tabular}{lc}
		\toprule
		Method & mIoU\\
		\midrule
		2D CNN (VGG16) & 57.2\\
		2D CNN (VGG19) & 58.3\\
		2D CNN (ResNet34) & 59.6\\
		\midrule
		2D CNN (VGG16) + PointNet++(SSG) & 65.0\\
		2D CNN (VGG19) + PointNet++(SSG) & 65.5\\
		2D CNN (ResNet34) + PointNet++(SSG) & 65.9\\
		\midrule
		2D CNN (VGG16) + PointNet++(MSG) & 65.0\\
		2D CNN (VGG16) + PointNet++(SSG, more centroids) & 66.4\\
		\bottomrule
	\end{tabular}
	\caption{2D CNN baselines and our MVPNet with different backbones. Results on validation set of ScanNetV2.}
	\label{tab:backbone}
\end{table}

To investigate the effect of stronger 2D backbones, we replace VGG16 with VGG19 and ResNet34.
Tab.~\ref{tab:backbone} shows the results of 2D CNN baselines and our MVPNet with different backbones on the validation set. Intuitively, stronger backbones lead to higher mIoU, and ResNet34 performs best. Due to runtime performance we choose VGG16 as backbone for ablation experiments and ResNet34 for best performance.

As to the stronger 3D backbone, we double the numbers of sampled centroids to (2048, 512, 128, 32), which increases the mIoU by 1.4. We also try to replace single-scale with multi-scale grouping (MSG) in PointNet++, but observe no improvement. As the image features already contain contextual information, it might not be necessary to process multiple scales explicitly with the MSG version.

\captionsetup[subfigure]{labelformat=empty}

\section{Conclusion}
While we can outperform state-of-the-art point based approaches by a significant margin using sliding window processing, methods that take the whole scene as input, e.g. the very well implemented SCN~\cite{graham20183d}, have a clear advantage.

We have proposed a framework to fuse 2D multi-view images and 3D point clouds in an effective way by computing image features in 2D first, lifting them to 3D, and then fuse complementary geometry and image information in canonical 3D space.
Comprehensive experiments are conducted on the ScanNetV2 Semantic Segmentation benchmark, which prove the advantage of calculating image features from multi-view images, and verify the superior robustness of our approach against voxel-based methods.

\section*{Acknowledgements}
We thank Li Yi and Wang Zhao to have provided us with the code of R-PointNet\cite{yi2018gspn}. We thank Valeo to have supported Maximilian Jaritz for his visit at UC San Diego.

{\small
\bibliographystyle{ieee}
\bibliography{egbib}
}

\appendix
\section{2D Encoder Decoder Architecture}
\begin{figure}[ht]
	\centering
	\includegraphics[width=\linewidth]{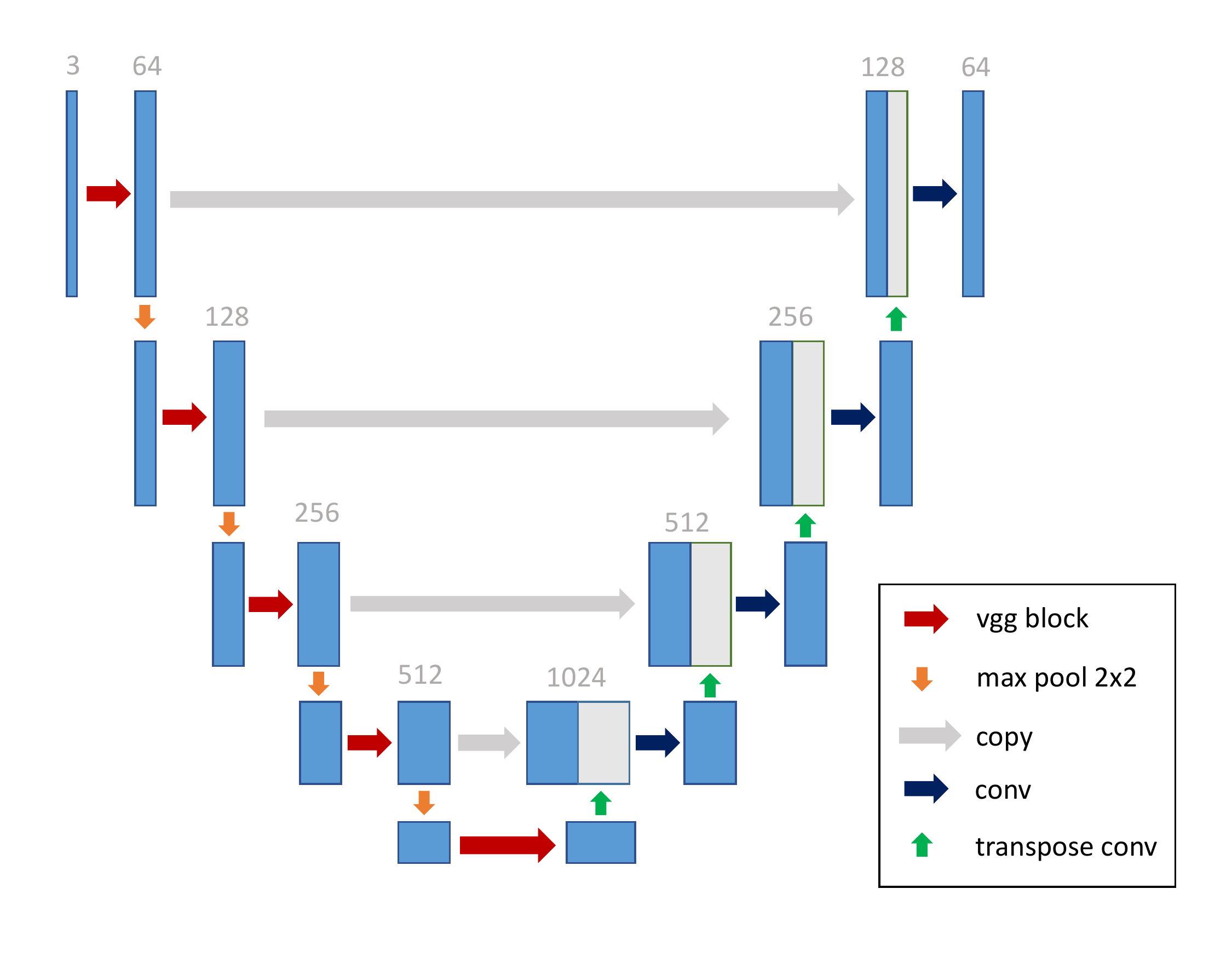}
	\caption{The architecture of the 2D encoder-decoder network.}
	\label{fig:archUnet}
\end{figure}
Fig.~\ref{fig:archUnet} illustrates the architecture of the 2D encoder-decoder network inspired by U-Net~\cite{ronneberger2015u}. We use VGG16 as encoder and initialize with ImageNet pre-trained weights. In the decoder, convolution is used to fuse concatenated features from skip connections, and transposed convolution for upsampling.

\section{Comparison with SparseConvNet (SCN)}
For lightweight SCN (small U-Net with 5cm-cubed voxels), we refer readers to the released codes\footnote{\href{https://github.com/facebookresearch/SparseConvNet/tree/master/examples/ScanNet}{github.com/facebookresearch/SparseConvNet}}.

\section{Experiments on S3DIS}
We evaluate on Stanford Indoor 3D (S3DIS) using images and xyx-maps from 2D-3D Semantics (2D-3D-S)~\cite{armeni2017joint}. Table~\ref{tab:s3dis} shows that MVPNet improves over previous methods by 4.16 mIoU.

\begin{table}[ht]
	\scriptsize
	\centering
	\begin{tabular}{lccc}
		\toprule
		Method & mIoU & mAcc & OA\\
		\midrule
		PointCNN\cite{li2018pointcnn} & 57.26 & 63.86 & 85.91 \\
		SPG\cite{landrieu2018large} & 58.04 & 66.50 & 86.38 \\
		PCCN\cite{wang2018deep} & 58.27 & 67.01 & - \\
		\midrule
		PointNet++\cite{qi2017pointnet++} (our implementation) & 56.19 & 64.09 & 85.26  \\
		MVPNet & \textbf{62.43} & \textbf{68.68} & \textbf{88.08} \\
		\bottomrule
	\end{tabular}
	\caption{Segmentation results on S3DIS Area 5.}
	\label{tab:s3dis}
\end{table}

\section{More Ablation Studies}
\begin{table}
	\scriptsize
	\centering
	\begin{tabular}{lc}
		\toprule
		Method & mIoU\\
		\midrule
		MVPNet(VGG19) & 66.6\\
		MVPNet(ResNet34) & 67.3\\
		\midrule
		MVPNet(ResNet34) + class weights & 68.0\\
		\midrule
		MVPNet(ResNet34) + ensemble & 68.3\\
		\bottomrule
	\end{tabular}
	\caption{The variants of MVPNet. The results are reported on the validation set of ScanNetV2.}
	\label{tab:variant}
\end{table}

For the ScanNetV2 3D semantic label benchmark, we employ MVPNet with 5 views and use ResNet34 as the 2D backbone. The numbers of centroids are 2048, 512, 128, 64 respectively.

Tab.~\ref{tab:variant} shows the comparison among several variants of the submission version. 
The stronger 2D backbone (ResNet34) improves the mIoU by 0.7 against the weaker 2D backbone (VGG19). 
Moreover, we also experiment with training MVPNet with class weights, which boosts the mIoU (+0.7) as the evaluation metric favors more balanced predictions.
To achieve the best performance (68.3), we ensemble 4 models of MVPNet with ResNet34.

\end{document}